\documentclass[10pt,journal,compsoc]{IEEEtran}
\usepackage{xspace}
\ifCLASSOPTIONcompsoc
\usepackage[nocompress]{cite}
\else
\usepackage{cite}
\fi
\usepackage{soul}
\usepackage{colortbl}
\usepackage{booktabs}
\usepackage{pifont}
\usepackage{subcaption}
\usepackage{multirow}
\usepackage{multicol}
\usepackage[ruled]{algorithm2e}
\usepackage{listings}
\usepackage{enumitem}

\ifCLASSINFOpdf
   \usepackage[pdftex]{graphicx}
\else
\fi
\usepackage{amsmath}
\usepackage{overpic}
\usepackage{url}
\usepackage[hidelinks]{hyperref}

\usepackage{todonotes}

\usepackage{soul}

\newcommand{\ourmethod}{CoSMix\xspace}

\definecolor{sourcecolor}{rgb}{0.78, 0.78, 0.78}
\definecolor{bestcolor}{rgb}{0.5, 0.95, 0.5}

\begin{document}

\title{Compositional Semantic Mix for Domain Adaptation in Point Cloud Segmentation}

\author{Cristiano~Saltori,
        Fabio~Galasso,
        Giuseppe~Fiameni,
        Nicu~Sebe,
        Fabio~Poiesi,
        and Elisa~Ricci
\IEEEcompsocitemizethanks{\IEEEcompsocthanksitem 
C. Saltori, N. Sebe and E. Ricci are with the Dept. of Information Engineering and Computer Science, University of Trento, Italy.
E-mails: cristiano.saltori@unitn.it, niculae.sebe@unitn.it, e.ricci@unitn.it.
\IEEEcompsocthanksitem F. Galasso is with the Dept. of Computer Science, Sapienza University of Rome, Italy.
\IEEEcompsocthanksitem G. Fiameni is with NVIDIA AI Technology Center, Italy.
\IEEEcompsocthanksitem F. Poiesi and E. Ricci are with Fondazione Bruno Kessler (FBK), Trento, Italy.
E-mail: poiesi@fbk.eu.\\

}
\thanks{Manuscript received April 19, 2005; revised August 26, 2015.}}

\markboth{Journal of \LaTeX\ Class Files,~Vol.~14, No.~8, August~2015}%
{Shell \MakeLowercase{\textit{et al.}}: Bare Demo of IEEEtran.cls for Computer Society Journals}

\IEEEtitleabstractindextext{%

\begin{abstract}
Deep-learning models for 3D point cloud semantic segmentation exhibit limited generalization capabilities when trained and tested on data captured with different sensors or in varying environments due to domain shift. 
Domain adaptation methods can be employed to mitigate this domain shift, for instance, by simulating sensor noise, developing domain-agnostic generators, or training point cloud completion networks. 
Often, these methods are tailored for range view maps or necessitate multi-modal input. 
In contrast, domain adaptation in the image domain can be executed through sample mixing, which emphasizes input data manipulation rather than employing distinct adaptation modules.
In this study, we introduce compositional semantic mixing for point cloud domain adaptation, representing the first unsupervised domain adaptation technique for point cloud segmentation based on semantic and geometric sample mixing. 
We present a two-branch symmetric network architecture capable of concurrently processing point clouds from a source domain (e.g.~synthetic) and point clouds from a target domain (e.g.~real-world). 
Each branch operates within one domain by integrating selected data fragments from the other domain and utilizing semantic information derived from source labels and target (pseudo) labels.
Additionally, our method can leverage a limited number of human point-level annotations (semi-supervised) to further enhance performance. 
We assess our approach in both synthetic-to-real and real-to-real scenarios using LiDAR datasets and demonstrate that it significantly outperforms state-of-the-art methods in both unsupervised and semi-supervised settings.
\end{abstract}

}

\maketitle

\IEEEdisplaynontitleabstractindextext


%
\IEEEpeerreviewmaketitle



\IEEEraisesectionheading{\section{Introduction}\label{sec:introduction}}

\IEEEPARstart{L}{IDAR} is currently the most suitable sensor for capturing accurate 3D measurements of an environment for autonomous driving~\cite{autoDrivingLevinson2011} and robotic navigation~\cite{loopZhou2022}. 
Semantic scene understanding is a crucial component for AI-based perception systems~\cite{lidar4autonomous}. 
LiDAR measurements can be analyzed in the form of 3D point clouds, with point cloud semantic segmentation used to assign a finite set of semantic labels to the 3D points~\cite{choy20194d}.
To train accurate deep learning models, large-scale datasets with point-level annotations are necessary~\cite{behley2019iccv, pan2020semanticposs, nuscenes2019}. 
This involves a costly and labor-intensive data collection process, as point clouds need to be captured in the real world and manually annotated. 
An alternative is to use synthetic data, which can be conveniently generated with simulators~\cite{csurka2017domain}. 
However, deep neural networks are known to suffer from domain shift when trained and tested on data from different domains~\cite{csurka2017domain}. 
Although simulators can reproduce the acquisition sensor with high fidelity, further research is still required to address such domain shift~\cite{synlidar}.

Data augmentation techniques based on the combination of samples and their labels, such as Mixup~\cite{mixup2018} or CutMix~\cite{Yun2019}, have been proposed to enhance deep network generalization. 
The underlying concept involves mixing samples to expand the training set and reduce overfitting. 
These methods were initially applied to image classification tasks and later adapted for domain adaptation and domain generalization in image recognition~\cite{xu2020adversarial, mancini2020towards}. 
Similar ideas have also been successfully extended to 2D semantic segmentation~\cite{tranheden2021dacs, gao2021dsp}.
While Unsupervised Domain Adaptation (UDA) for semantic segmentation in the image domain has been extensively studied~\cite{wang2021domain, gao2021dsp, tranheden2021dacs, zou2018unsupervised, zou2019confidence}, less attention has been devoted to developing adaptation techniques for point cloud segmentation. 
Point cloud UDA can be addressed in the input space~\cite{synlidar, zhao2020epointda} with dropout rendering~\cite{zhao2020epointda} or adversarial networks~\cite{synlidar}, or in the feature space through feature alignment~\cite{wu2019squeezesegv2}. 
A few studies have proposed exploiting sample mixing for point cloud data~\cite{Nekrasov213DV,Zou2021}, but they target different applications than UDA for semantic segmentation.

In this paper, we present a novel domain adaptation framework for 3D point cloud segmentation, named \ourmethod, which extends the approach presented in~\cite{saltori2022cosmix} to the semi-supervised settings (SSDA).
\ourmethod is designed to mitigate the domain shift by mixing semantically-informed groups of points (patches) across domains.
Specifically, we design a two-branch symmetric deep neural network pipeline that concurrently processes point clouds from a source domain (e.g.~synthetic or real) and point clouds from a target domain (e.g.~real or real but captured with a different sensor).
Target point clouds can be either unlabeled or partially labeled if one wants to use \ourmethod for UDA or SSDA, respectively.
Each branch is domain specific, i.e.~the source branch is in charge of mixing a source point cloud with selected patches of a target point cloud, and vice versa for the target branch.
We formulate mixing as a composition operation, which is similar to the concatenation operation proposed in \cite{Nekrasov213DV, Zou2021}, but unlike them, we leverage the semantic information to mix domains.
Patches from the source point cloud are selected based on the semantic labels of their points.
Patches from the target point cloud can be selected based on the predicted semantic pseudo-labels in the case of UDA and based on human annotations in the case of SSDA.
We will show that only a handful of manually annotated points are sufficient to significantly improve the domain adaptation performance.
When patches are mixed across domains we apply data augmentation both at local and global semantic levels to boost the efficacy of the mixing.
An additional key difference between our method and \cite{Nekrasov213DV, Zou2021} is the teacher-student learning scheme that we implement to improve the accuracy of the pseudo-labels.
We evaluate \ourmethod on large scale point cloud segmentation benchmarks, featuring both synthetic and real-world data, in several directions such as synthetic to real and real to real.
Specifically, we use the following datasets: 
SynLiDAR~\cite{synlidar}, 
SemanticPOSS~\cite{pan2020semanticposs}, 
SemanticKITTI~\cite{behley2019iccv}, and
nuScenes~\cite{nuscenes2019}.
Our results show that \ourmethod can reduce the domain shift, outperforming state-of-the-art methods in both UDA and SSDA settings.
We perform detailed analyses of \ourmethod and an ablation study of each component, highlighting its strengths and discussing its limitations.

This paper extends our earlier work \cite{saltori2022cosmix} in several aspects.
We extend the original CoSMix in order to tackle the SSDA setup.
The current design allows a user to input a few annotated points to significantly improve the semantic segmentation performance on the target domain.
Then, we significantly extend our experimental evaluation and analysis by adding new experiments, new comparisons, and new ablation studies to evaluate this new setup.
We extend the related work by thoroughly reviewing additional state-of-the-art approaches, and summarizing these approaches in a comprehensive table that highlights key contributions and setups.
Lastly, the code is available at \url{https://github.com/saltoricristiano/cosmix-uda}.




\section{Related work}\label{sec:related}

\begin{table*}[t]
    \centering
    \caption{
    Overview of existing methods for unsupervised (UDA) and semi-supervised (SSDA) adaptation in point cloud segmentation. 
    For each approach, we report the sensor setup (Setup), the architecture (Input data type and Model), and the source and target datasets. 
    Then, we classify the adaptation strategy into mixup based, adversarial learning based, alignment based, generative based, self-training based and auxiliary task based. Furthermore, we report whether the implementation (Code) is publicly available.}
    \vspace{-1mm}
    \label{tab:review_related}
    \tabcolsep 4pt
    \resizebox{\textwidth}{!}{%
    \begin{tabular}{l|c|cc|cc|cc|cccccc|c}
        \toprule
        \multirow{2}{*}{\textbf{Method}} & \multirow{2}{*}{\textbf{Setup}} & \multicolumn{2}{c|}{\textbf{Architecture}} & \multicolumn{2}{c|}{\textbf{Datasets}} & \multicolumn{2}{c|}{\textbf{Settings}} & \multicolumn{6}{c|}{\textbf{Adaptation}} & \multirow{2}{*}{\textbf{Code}}\\
         & & \textbf{Input data} & \textbf{Model} & \textbf{Source} & \textbf{Target} & \textbf{UDA} & \textbf{SSDA} & \textbf{Mixup} &  \textbf{Adv.} & \textbf{Align.} & \textbf{Gen.} & \textbf{Self-train.} & \textbf{Aux. task} &\\
        \midrule
        RayCast~\cite{langer2020domain} & real-to-real & RV & RangeNet++~\cite{milioto2019rangenet++} & Sem.KITTI~\cite{behley2019iccv} & nuSc.~\cite{nuscenes2019} & \ding{51} & & & & \ding{51} & \ding{51} & & \\
        
         & & & & & & & & & & & & & & \\
         
        \multirow{2}{*}{ePointDA~\cite{zhao2020epointda}} & \multirow{2}{*}{synth-to-real} & \multirow{2}{*}{RV} & \multirow{2}{*}{SqueezeSegV2~\cite{wu2019squeezesegv2}} & \multirow{2}{*}{GTA-V~\cite{wu2019squeezesegv2}} & KITTI~\cite{geiger2013vision} & \multirow{2}{*}{\ding{51}} & & & \multirow{2}{*}{\ding{51}} & \multirow{2}{*}{\ding{51}} & \multirow{2}{*}{\ding{51}} & & & \\
         & & & & & Sem.KITTI~\cite{behley2019iccv} & & & & & & & & & \\
        
         & & & & & & & & & & & & & & \\
        
        SqueezeSegV2~\cite{wu2019squeezesegv2} & synth-to-real & RV & SqueezeSegV2~\cite{wu2019squeezesegv2} & GTA-V~\cite{wu2019squeezesegv2} & KITTI~\cite{geiger2013vision} & \ding{51} & & & & \ding{51} & & & & \ding{51}\\
        
         & & & & & & & & & & & & & & \\

        \multirow{3}{*}{Gated~\cite{rochan2021unsupervised}} & synth-to-real & \multirow{3}{*}{RV} & \multirow{3}{*}{SalsaNext~\cite{cortinhal2020salsanext}} & GTA-V~\cite{wu2019squeezesegv2} & \multirow{2}{*}{KITTI~\cite{geiger2013vision}} & \multirow{3}{*}{\ding{51}} & & & & & \multirow{3}{*}{\ding{51}} & & \multirow{3}{*}{\ding{51}} & \\
         & \multirow{2}{*}{real-to-real} &  &  & nuScenes~\cite{nuscenes2019} & & & & & & & & & & \\
          &  &  &  & KITTI~\cite{geiger2013vision} & nuScenes~\cite{nuscenes2019}  & & & & & & & & & \\
        
        \midrule
        \multirow{3}{*}{xMUDA~\cite{jaritz2019xmuda}}& \multirow{3}{*}{real-to-real} & \multirow{3}{*}{2D\&3D} & \multirow{3}{*}{xMUDA~\cite{jaritz2019xmuda}} & nuSc.~\cite{nuscenes2019} & nuSc.\cite{nuscenes2019} & \multirow{3}{*}{\ding{51}} & & & &\multirow{3}{*}{\ding{51}} & & \multirow{3}{*}{\ding{51}} & & \multirow{3}{*}{\ding{51}}\\
         & & & & KITTI~\cite{geiger2012we} & KITTI~\cite{geiger2012we} & & & & & & & & & \\
          & & & & A2D2~\cite{geyer2020a2d2} & A2D2~\cite{geyer2020a2d2} & & & & & & & & & \\
          
         & & & & & & & & & & & & & & \\
        
        \multirow{5}{*}{Cross-modal~\cite{jaritz2022cross}}& & \multirow{5}{*}{2D\&3D} & \multirow{5}{*}{xMUDA~\cite{jaritz2019xmuda}} & nuSc.~\cite{nuscenes2019} &  & \multirow{5}{*}{\ding{51}} & \multirow{5}{*}{\ding{51}} & & & \multirow{5}{*}{\ding{51}} & & \multirow{5}{*}{\ding{51}} & & \multirow{5}{*}{\ding{51}}\\
         & real-to-real & & & v.KITTI~\cite{cabon2020virtual} & nuSc.~\cite{nuscenes2019} & & & & & & & & & \\
         & synth-to-real & & & A2D2~\cite{geyer2020a2d2} & Sem.KITTI~\cite{behley2019iccv} & & & & & & & & & \\
         & & & & Sem.KITTI~\cite{behley2019iccv} & Waymo~\cite{sun2020scalability} & & & & & & & & & \\
         & & & & Waymo~\cite{sun2020scalability} &  & & & & & & & & & \\
        
        \midrule
        \multirow{3}{*}{Complete\&Label~\cite{yi2021complete}} & \multirow{3}{*}{real-to-real} & \multirow{3}{*}{3D} & \multirow{3}{*}{SparseConv~\cite{SubmanifoldSparseConvNet}} &  KITTI~\cite{geiger2013vision} & KITTI~\cite{geiger2013vision} & \multirow{3}{*}{\ding{51}} & & & & & \multirow{3}{*}{\ding{51}} & & &\\
         & & & & Waymo~\cite{sun2020scalability} & Waymo~\cite{sun2020scalability} & & & & & & & & &\\
         & & & & nuSc.~\cite{nuscenes2019} & nuSc.~\cite{nuscenes2019} & & & & & & & & &\\
         
         & & & & & & & & & & & & & & \\
         
        \multirow{2}{*}{PCT~\cite{synlidar}} & \multirow{2}{*}{synth-to-real} & \multirow{2}{*}{3D} & \multirow{2}{*}{Minkowski~\cite{choy20194d}} & \multirow{2}{*}{SynLiDAR~\cite{synlidar}} & Sem.KITTI~\cite{behley2019iccv} & \multirow{2}{*}{\ding{51}} & \multirow{2}{*}{\ding{51}} & & & & \multirow{2}{*}{\ding{51}} & & & \multirow{2}{*}{\ding{51}} \\
         & & & & & Sem.POSS~\cite{pan2020semanticposs} & & & & & & & & &\\
         
         & & & & & & & & & & & & & & \\

        \multirow{2}{*}{\ourmethod~\cite{saltori2022cosmix}}  & \multirow{2}{*}{synth-to-real} & \multirow{2}{*}{3D} & \multirow{2}{*}{Minkowski~\cite{choy20194d}} & \multirow{2}{*}{SynLiDAR~\cite{synlidar}} & Sem.KITTI~\cite{behley2019iccv} & \multirow{2}{*}{\ding{51}} & & \multirow{2}{*}{\ding{51}} & & & & \multirow{2}{*}{\ding{51}} & & \multirow{2}{*}{\ding{51}}\\
         & & & & & Sem.POSS~\cite{pan2020semanticposs} & & & & & & & & &\\
         
         & & & & & & & & & & & & & & \\
         
         \multirow{3}{*}{Ours}  & real-to-real & \multirow{3}{*}{3D} & \multirow{3}{*}{Minkowski~\cite{choy20194d}} & SynLiDAR~\cite{synlidar} & Sem.KITTI~\cite{behley2019iccv} & \multirow{3}{*}{\ding{51}} & \multirow{3}{*}{\ding{51}} & \multirow{3}{*}{\ding{51}} & & & & \multirow{3}{*}{\ding{51}} & & \multirow{3}{*}{\ding{51}}\\
         & synth-to-real & & & SemKITTI~\cite{behley2019iccv} & Sem.POSS~\cite{pan2020semanticposs} & & & & & & & & &\\
         & & & & & nuScenes~\cite{nuscenes2019} & & & & & & & & &\\
        \bottomrule
    \end{tabular}
    }
\end{table*}

\noindent\textbf{Point cloud semantic segmentation.}
Point cloud semantic segmentation can be performed at point level \cite{qi2017pointnet++}, on range views \cite{ronneberger2015u} or on a voxelized point clouds \cite{zhou2018voxelnet}.
Point-level architectures process the input point cloud without the need for intermediate representation processing.
This architectures include PointNet~\cite{qi2017pointnet}, which is based on a series of multilayer perceptrons.
PointNet++~\cite{qi2017pointnet++} improves on PointNet by aggregating global and local point features at multiple scales.
RandLA-Net~\cite{hu2020randla} extends PoinNet++ \cite{qi2017pointnet++} by embedding local spatial encoding, random sampling and attentive pooling.
KPConv~\cite{thomas2019kpconv} learns weights in the continuous space, and introduces flexible and deformable convolutions for point cloud processing.
These methods are computationally inefficient when large-scale point clouds are processed.
Computational efficiency can be improved by projecting 3D points on 2D representations \cite{milioto2019rangenet++} or by using 3D quantization approaches \cite{choy20194d}.
The former includes 2D projection-based approaches that use 2D range maps and exploit standard 2D convolution filters~\cite{ronneberger2015u} to segment these maps prior to a re-projection in the 3D space. 
RangeNet++~\cite{milioto2019rangenet++}, SqueezeSeg networks~\cite{wu2018squeezeseg, wu2019squeezesegv2}, 3D-MiniNet~\cite{alonso2020MiniNet3D} and PolarNet~\cite{zhang2020polarnet} are approaches that belong to this category.
Although these approaches are efficient, they tend to lose information when the input data are projected in 2D and re-projected in 3D.
The latter includes 3D quantization-based approaches that transform the input point cloud into a 3D discrete representations, and that employ 3D convolutions~\cite{zhou2018voxelnet} or 3D sparse convolutions~\cite{SubmanifoldSparseConvNet, choy20194d} to predict per-point classes.
VoxelNet~\cite{zhou2018voxelnet} maps input points into a voxel-grid and processes the input voxel-grid with 3D convolutions.
SparseConv~\cite{SubmanifoldSparseConvNet, 3DSemanticSegmentationWithSubmanifoldSparseConvNet} and MinkowskiNet~\cite{choy20194d} improves voxel processing and introduce sparse convolutions to improve efficiency.
Cylinder3D~\cite{zhu2021cylindrical} further improves voxel processing for LiDAR data by using cylindrical and asymmetrical 3D convolutions.
In our work, we use MinkowskiNet~\cite{choy20194d}, which provides a trade off between accuracy and efficiency.

\vspace{1mm}
\noindent\textbf{Sample mixing for 2D domain adaptation.}
Unsupervised domain adaptation (UDA) for image-based tasks is a well-studied problem~\cite{toldo2020unsupervised}, and there exist several methods using domain adversarial learning~\cite{tzeng2017adversarial, vu2019advent, hoffman2018cycada}, regularization losses~\cite{roy2019unsupervised, vu2019advent, roy2019unsupervised}, self-training~\cite{zou2018unsupervised, pilzer2019refine}, multi-task learning~\cite{vu2019dada}, and curriculum learning~\cite{zhang2019curriculum}. 
Mixup can also be used for domain adaptation~\cite{mixup2018, Yun2019, olsson2021classmix}.
For 2D semantic segmentation, several recent works have used sample mixing for domain adaptation, including BAPA-Net~\cite{liu2021bapa}, CAMix~\cite{zhou2022context}, DACS~\cite{tranheden2021dacs}, DSP~\cite{gao2021dsp}, and DAFormer~\cite{hoyer2022daformer}. BAPA-Net employs a boundary-informed mixing strategy, while CAMix proposes a context-aware mask generation for domain mixing. 
DACS extends ClassMix~\cite{olsson2021classmix} for domain adaptation by pasting specific source classes into target images. 
DSP improves on DACS by introducing self-training, soft-labels, and a double mixing strategy. 
Unlike these works, our method tackles UDA for 3D semantic segmentation. 
It should be noted that extending image mixing strategies to point clouds is not as straightforward, as point clouds are sparse in nature. 
Therefore, we introduce a novel mixing strategy specifically designed for 3D point clouds.

\vspace{1mm}
\noindent\textbf{Domain adaptation for point cloud segmentation.}
Unlike domain adaptation for image-based tasks~\cite{toldo2020unsupervised, kouw2021review}, domain adaptation for point cloud segmentation still lacks a unified experimental setup to compare different approaches.
We review domain adaptation approaches for point cloud segmentation by grouping them into range-view methods, multi-modal (2D\&3D) methods, and 3D-focused methods.
Tab.~\ref{tab:review_related} provides a detailed summary of these approaches.

\noindent\textit{Range-view} (RV) images are computed through a cylindrical projection of the input point cloud onto a 2D plane.
After projection, RV images can be processed with existing 2D convolution networks.
RV-based networks are affected by domain shift, which can be mitigated by using generative approaches~\cite{langer2020domain, zhao2020epointda}, feature alignment~\cite{langer2020domain, wu2019squeezesegv2, zhao2020epointda}, and contrastive learning~\cite{rochan2021unsupervised}.
RayCast~\cite{langer2020domain} tackles the real-to-real UDA problem by transferring the sensor pattern of the target domain to the source domain through ray casting.
After training the deep network on the source data, a minimal-entropy correlation alignment loss is used to reduce domain shift~\cite{morerio2017minimal}.
SqueezeSegV2~\cite{wu2019squeezesegv2} improves the SqueezeSeg~\cite{wu2018squeezeseg} architecture, and reduces domain shift in the synth-to-real setup by aligning source and target features with a geodesic correlation alignment~\cite{morerio2017minimal}.
ePointDA~\cite{zhao2020epointda} addresses domain shift in the synth-to-real UDA setup at both input level and feature level. 
At input level, a generative CycleGAN~\cite{CycleGAN2017} is trained to simulate real sensor noise on synthetic source data. 
At feature level, a higher-order momentum loss~\cite{chen2020homm} is used to learn domain agnostic features between source and target input data.
Gated~\cite{rochan2021unsupervised} states that domain shift between source and target point clouds can be mitigated by solving the sparsity shift and by introducing domain specific parameters. 
Given an input pair of source an target RV images, they first solve the sparsity difference thorough self-supervised completion and by applying a dropout mask. 
Then, residual gated adapters are added to the segmentation model to learn target specific parameters.
None of the RV-based methods tackle the semi-supervised scenario.

\noindent\textit{Multi-modal models} are designed to process the information captured by multiple input sensors, e.g.~RGB cameras and LiDAR sensors are those typically used.
Domain shift is tackled by enforcing prediction consistency among modalities and domains, and by using target (pseudo) labels.
xMUDA~\cite{jaritz2019xmuda} uses cross-modality and cross-domain consistency to learn a domain agnostic model in the real-to-real UDA setup. 
Cross-modal consistency exploits source labels and target pseudo-labels to produce consistent multi-modal predictions in both the domains. 
DeepCORAL feature alignment~\cite{sun2016deep} is used to enforce feature alignment between source and target domains.
In~\cite{jaritz2022cross}, xMUDA is extended to SSDA settings showing that cross-modal consistency is effective even in the semi-supervised settings.

\noindent\textit{3D methods} can process input point clouds with or without prior voxelization.
UDA approaches for 3D segmentation include voxel-based architectures such as SparseConv~\cite{3DSemanticSegmentationWithSubmanifoldSparseConvNet} and MinkowskiNet~\cite{choy20194d}. 
Domain shift can be tackled by focusing on the problem of sparsity \cite{yi2021complete, synlidar} or by employing mix up strategies \cite{saltori2022cosmix}.
Complete\&Label~\cite{yi2021complete} reduces the sparsity difference between real domains by formulating the domain adaptation problem as a point cloud completion (or densification) problem. 
A self-supervised completion network is trained to make the sparse input point cloud denser.
The pre-processed point clouds can then be used as intermediate domains in order to lower the domain shift.
PCT~\cite{synlidar} disentangles domain shift between synthetic and real point clouds into appearance and sparsity.
Then, PCT learns an appearance translation module and a sparsity translation module.
These modules are used for translating source data in the target modality. 
Translated data are then used together with ST~\cite{zou2019confidence} and APE~\cite{kim2020attract} in the UDA and SSDA settings, respectively.

\ourmethod~\cite{saltori2022cosmix} is a method that reduces domain shift in point cloud data by introducing a compositional semantic mixup strategy with a teacher-student learning scheme. 
The method obtains domain-invariant models/features by creating two new intermediate domains of composite point clouds: a mixed source and a mixed target. In the mixed target, source instances pull the target domain closer to the source domain, preventing overfitting from noisy pseudo-labels. 
In the mixed source, target instances (pseudo-labels) bring the target modality into the source domain, pulling the source domain closer to the target domain. 
The teacher-student learning scheme enables the iterative improvement of pseudo-labels, progressively reducing the domain gap.
In this work, we extend CoSMix~\cite{saltori2022cosmix} to the SSDA settings by allowing target labels to be mixed in the source and target (unlabeled) point clouds while improving adaptation. 
We also show how a small amount of target supervision can significantly improve the adaptation performance.



\section{Compositional Semantic Mix (\ourmethod)}\label{sec:method_uda}

\subsection{Preliminaries and definitions}

\begin{figure*}[t]
    \centering
    \includegraphics[width=\textwidth]{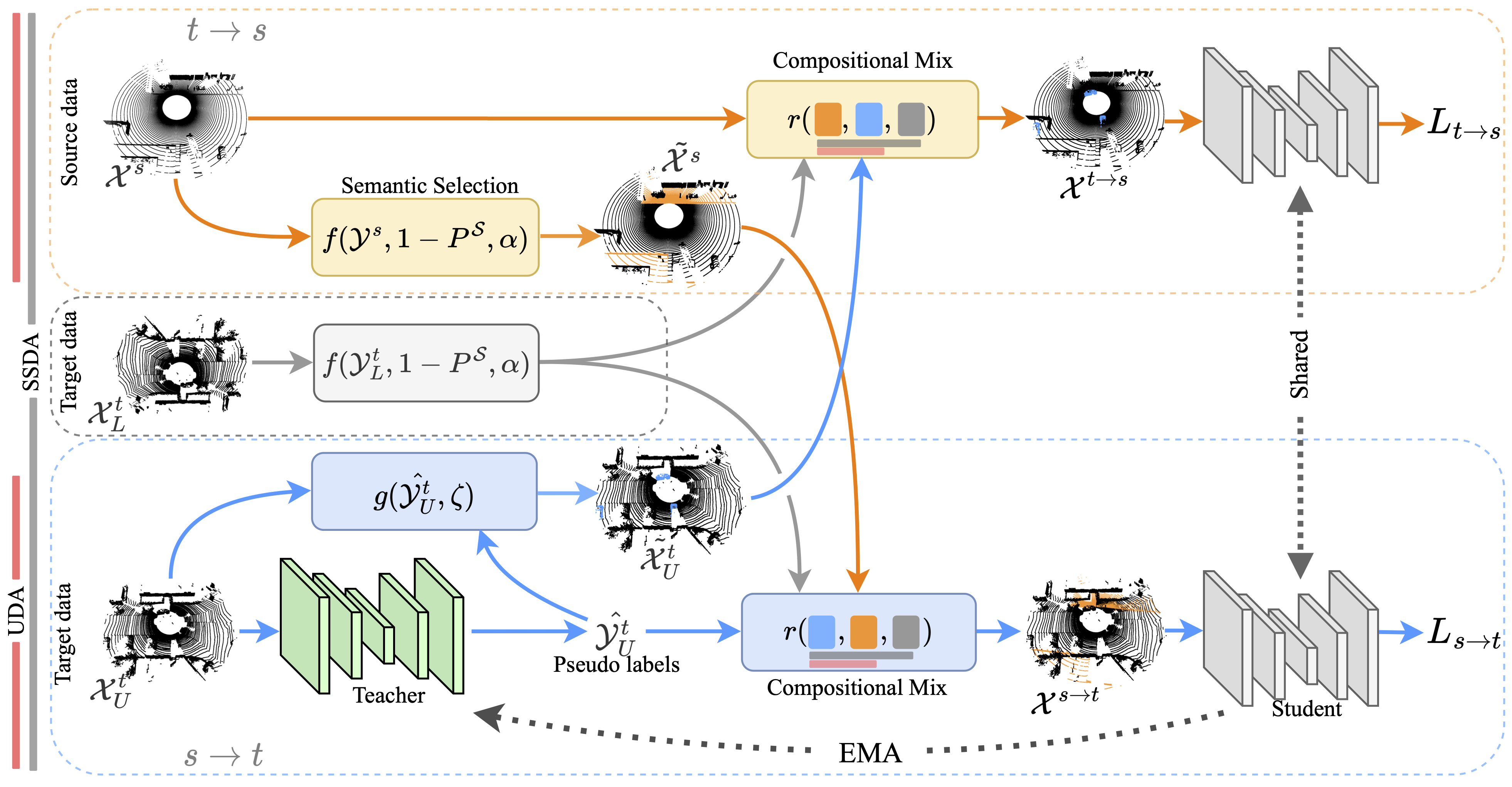}
    \vspace{-6mm}
    \caption{Block diagram of \ourmethod detailing the UDA and SSDA settings. 
    The UDA setting uses the top and bottom branch (red line). 
    The SSDA setting uses also the middle branch in addition to those used in UDA (gray line). 
    In the top branch, the input source point cloud $\mathcal{X}^s$ is mixed with the unsupervised target point cloud $\mathcal{X}^t_\textsf{U}$ obtaining $\mathcal{X}^{t\rightarrow s}$. In the bottom branch, the input target point cloud $\mathcal{X}^t_\textsf{U}$ is mixed with the source point cloud $\mathcal{X}^s$ obtaining $\mathcal{X}^{s\rightarrow t}$.
    In the SSDA setting, the labeled target data $\mathcal{X}^t_\textsf{L}$ are mixed with the source point cloud $\mathcal{X}^s$ and with the unsupervised target point cloud $\mathcal{X}^t_\textsf{U}$.
    A teacher-student learning architecture is used in both the UDA and SSDA settings to improve pseudo-label accuracy while adapting over target domain. 
    This is achieved by updating the teacher network through Exponential Moving Average (EMA).
    Semantic Selection ($f$ and $g$) selects subsets of points (patches) to be mixed based on the source labels $\mathcal{Y}^s$, the target labels $\mathcal{Y}^t_\textsf{L}$, and target pseudo-labels $\hat{\mathcal{Y}^t_\textsf{U}}$ information. 
    Compositional Mix applies local $h$ and global $r$ augmentations and mixes the selected patches among domains.}
    \label{fig:block_diagram}
\end{figure*}

\ourmethod implements a teacher-student learning scheme that exploits the supervision from the source domain, the self-supervision from the target domain and, if available, the supervision from a few labeled target samples to improve the semantic segmentation on the target domain.
Our method is trained on two different mixed point cloud sets.
The first is the composition of the source point cloud with pseudo-labeled portions of points, or \textit{patches}, of the unlabeled target point cloud. 
Target patches bring the target modality in the source domain making the altered source domain more similar to the target domain.
The second is the composition of the unlabeled target point cloud with randomly selected patches of the source point cloud. 
Source patches make the altered target domain more similar to the source domain, preventing overfitting from noisy pseudo-labels.
If available, labeled points of the target point clouds can also be used in both the mixed point cloud sets.
This target supervision can further reduce domain shift.
The teacher-student learning scheme iteratively improves pseudo labels, progressively reducing the domain gap.
Fig.~\ref{fig:block_diagram} illustrates the block diagram of \ourmethod.

Let $\mathcal{S} = \{(\mathcal{X}^s,\mathcal{Y}^s)\}$ be the source dataset that is composed of $N^s = |\mathcal{S}|$ labeled point clouds, where $\mathcal{X}^s$ is a point cloud and $\mathcal{Y}^s$ is its point-level labels, and $|.|$ is the cardinality of a set.
Labels take values from a set of semantic classes $\mathcal{C} = \{ c \}$, where $c$ is a semantic class.
Let $\mathcal{T}_\textsf{U} = \{\mathcal{X}^t_\textsf{U}\}$ be the unlabeled target dataset composed of $N^t_\textsf{U} = |\mathcal{T}_\textsf{U}|$ unlabeled point clouds.
Let $\mathcal{T}_\textsf{L} = \{(\mathcal{X}^t_\textsf{L}, \mathcal{Y}^t_\textsf{L})\}$ be the semi-supervised set of $N^t_\textsf{L} = |\mathcal{T}_\textsf{L}|$ labeled target point clouds with $N^t_\textsf{L} \ll N^t_\textsf{U}$.

On the upper branch, the source point cloud $\mathcal{X}^s$ is mixed with selected patches of the target point cloud $\mathcal{X}^t_\textsf{U}$ and, selected patches of the supervised point cloud $\mathcal{X}^t_\textsf{L}$ when available.
The unlabeled target patches from $\mathcal{X}^t_\textsf{U}$ are subsets of points that correspond to the most confident pseudo-labels $\hat{\mathcal{Y}^t_\textsf{U}}$ that the teacher network produces during training.
The supervised target patches are subsets of points that are randomly selected based on the class frequency distribution in the source training set.
On the lower branch, the target point cloud $\mathcal{X}^t_\textsf{U}$ is mixed with the selected patches of the source point cloud $\mathcal{X}^s$ and with the selected patches of $\mathcal{X}^t_\textsf{L}$, if available.
The source patches are subsets of points that are randomly selected based on their class frequency distribution in the training set.

We define the branch that mixes target point cloud patches to the source point cloud as $t \rightarrow s$ and the branch that does the vice versa as $s \rightarrow t$.
Let $\mathcal{X}^{t \rightarrow s}$ be the mixed point cloud obtained from the upper branch, and $\mathcal{X}^{s \rightarrow t}$ be the mixed point cloud obtained from the lower branch.
Lastly, let $\Phi_\mathcal{\theta}$ and $\Phi_\mathcal{\theta'}$ be the student and teacher deep networks with learnable parameters $\theta$ and $\theta'$, respectively.

\subsection{Semantic selection}\label{sec:semantic_selection}

To train the student networks with balanced data, we perform a selection of reliable and informative point cloud patches prior to mixing points and labels across domains.
To select patches from the source point cloud, we use the class frequency distribution by counting the number of points of each semantic class within $\mathcal{S}$.
Unlike DSP~\cite{gao2021dsp} that selects long-tail classes in advance, we exploit the source distribution and the semantic classes available to dynamically sample classes at each iteration.

Let $P_\mathcal{Y}^s$ be the class frequency distribution of $\mathcal{S}$.
We create a function $f$ that randomly selects a subset of classes at each iteration based on the labels $\tilde{\mathcal{Y}}^s \subset \mathcal{Y}^s$.
$f$ performs a weighted random sampling of $\alpha$ classes from the input point cloud by using $1-P_\mathcal{Y}^s$ as the class weight for each class. 
$\alpha$ is an hyperparameter that regulates the ratio of selected classes for each point cloud.
The output of $f$ is a set point-level labels belonging to the sampled classes, i.e.~$\tilde{\mathcal{Y}}^s$.
The likelihood that $f$ selects a class $c$ is inversely proportional to its class frequency in $\mathcal{S}$.
Formally we have 
\begin{equation}
    \tilde{\mathcal{Y}}^s = f(\mathcal{Y}^s, 1-P_\mathcal{Y}^s, \alpha).
    \label{eq:source_sampling}
\end{equation}
Example: with $\alpha=0.5$, the algorithm selects a number of patches corresponding to the 50\% of the available classes, i.e.~long-tailed classes are selected with a higher likelihood.

Let $\tilde{\mathcal{X}}^s$ be the set of points that correspond to $\tilde{\mathcal{Y}}^s$, and let $\tilde{\mathcal{X}}^s_c \subset \tilde{\mathcal{X}}^s$ be a patch (set of points) that belongs to class $c \in \mathcal{C}$.
To select patches from the target point clouds, we apply the same set of operations but using the pseudo-labels produced by the teacher network based on their prediction confidence.
Specifically, we define a function $g$ that selects reliable pseudo-labels based on their confidence value. 
The selected pseudo-labels are defined as 
\begin{equation}\label{eq:target_pseudo}
    \tilde{\mathcal{Y}}^t_\textsf{U} = g(\Phi_{\theta'}(\mathcal{X}^t_\textsf{U}), \zeta),
\end{equation}
where $\Phi_{\theta'}$ is the teacher network, $\zeta$ is the confidence threshold used by the function $g$ and $\tilde{\mathcal{Y}}^t_\textsf{U} \subset \hat{\mathcal{Y}}^t_\textsf{U}$.

Let $\tilde{\mathcal{X}}^t_\textsf{U}$ be the set of points that correspond to $\tilde{\mathcal{Y}}^t_\textsf{U}$.

In the case of target supervision, we apply $f$ to the target labels $\mathcal{Y}^t_\textsf{L}$ and randomly select target patches as
\begin{equation}
    \tilde{\mathcal{Y}}^t_\textsf{L} = f(\mathcal{Y}^t_\textsf{L}, 1-P_\mathcal{Y}^s, \mu),
    \label{eq:target_sampling}
\end{equation}
where $\mu$ is an hyperparameter that regulates the ratio of selected classes for each point cloud similarly to $\alpha$.

\subsection{Compositional mix}\label{sec:compositional_mix}

The goal of our compositional mixing module is to create mixed point clouds based on the selected semantic patches.
The compositional mix involves three consecutive operations:
\textit{local random augmentation}, where patches are augmented randomly and independently from each other;
\textit{concatenation}, where the augmented patches are concatenated to the point cloud of the other domain to create the mixed point cloud;
\textit{global random augmentation}, where the mixed point cloud is randomly augmented.
This module is applied twice, once for the $t \rightarrow s$ branch (top of Fig.~\ref{fig:block_diagram}), where target patches are mixed within the source point cloud, and once for the $s \rightarrow t$ branch (bottom of Fig.~\ref{fig:block_diagram}), where source patches are mixed within the target point cloud.
Unlike Mix3D~\cite{Nekrasov213DV}, our mixing strategy embeds data augmentation at local level and global level.

Let $\delta$ be the indicator function that we define as
\begin{equation}
    \delta(\mathcal{T}_\textsf{L}) = 
    \begin{cases}
        1 \,\,\,\,\,\,\, if \,\,\,\,\, \mathcal{T}_\textsf{L}\neq \emptyset\\
        0 \,\,\,\,\,\,\, \text{otherwise},
    \end{cases}
\end{equation}
which indicates whether the supervised target set $\mathcal{T}_\textsf{L}$ is empty or not. 
This can be interpreted as the user desire or need to use additional target supervision.

In the $s \rightarrow t$ branch, we apply the local random augmentation $h$ to all the points $\tilde{\mathcal{X}}^s_c \subset \tilde{\mathcal{X}}^s$.
We repeat this operation for all $c \in \tilde{\mathcal{Y}}^s$.
Note that $h$ is a local and random augmentation that produces a different result each time it is applied to a set of points.
We define the result of this operation as
\begin{equation}\label{eq:local_aug_source}
    h(\tilde{\mathcal{X}}^s) = \left \{ h(\tilde{\mathcal{X}}^s_c), \forall c \in \tilde{\mathcal{Y}}^s \right \}.
\end{equation}

If $\delta(\mathcal{T}_\textsf{L}) = 1$ we can apply $h$ also to $\tilde{\mathcal{X}}^t_\textsf{L}$ and obtain
\begin{equation}\label{eq:local_aug_target}
    h(\tilde{\mathcal{X}}^t_\textsf{L}) = \left \{ h(\tilde{\mathcal{X}}^t_{\textsf{L}}), \forall c \in \tilde{\mathcal{Y}}^t_\textsf{L} \right \}.
\end{equation}

Then, we concatenate the locally augmented patches with the target point cloud $X^t_\textsf{L}$ and we apply the global random augmentation, such as
\begin{equation}\label{eq:point_mix}
    \mathcal{X}^{s \rightarrow t} = \begin{cases}
    r(h(\tilde{\mathcal{X}}^s) \cup h(\tilde{\mathcal{X}}^t_\textsf{L}) \cup \mathcal{X}^t_\textsf{U}) \,\,\,\,\, if \,\,\,\, \delta(\mathcal{T}_\textsf{L}) = 1,\\\\
    r(h(\tilde{\mathcal{X}}^s) \cup \mathcal{X}^t_\textsf{U}) \,\,\,\,\,\,\,\,\,\,\,\, \text{otherwise}
    \end{cases}
\end{equation}
Their respective labels are concatenated accordingly as
\begin{equation}\label{eq:label_mix}
    \mathcal{Y}^{s \rightarrow t} = \begin{cases}
    r(h(\tilde{\mathcal{Y}}^s) \cup h(\tilde{\mathcal{Y}}^t_\textsf{L}) \cup \mathcal{Y}^t_\textsf{U}) \,\,\,\,\, if \,\,\,\, \delta(\mathcal{T}_\textsf{L}) = 1,\\\\
    r(h(\tilde{\mathcal{Y}}^s) \cup \mathcal{Y}^t_\textsf{U}) \,\,\,\,\,\,\,\,\,\,\,\, \text{otherwise},
    \end{cases}
\end{equation}
where $r$ is the global augmentation function.

The same operations of Eq.~\ref{eq:point_mix}-\ref{eq:label_mix} are also performed in the $t \rightarrow s$ branch by mixing target patches within the source point cloud. Instead of using source labels, we use the teacher network to generate pseudo-labels from the target data. 
Additionally, we use target supervision if $\delta(\mathcal{T}_\textsf{L})=1$.
Then, we concatenate them with the labels of the source data.
This results in $\mathcal{X}^{t \rightarrow s}$ and $\mathcal{Y}^{t \rightarrow s}$.
Note that $\mathcal{T}_\textsf{L}$ may be used without compositional mix and without double branched mixing. 
We implement $h$ and $r$ by using typical augmentation strategies for point clouds~\cite{choy20194d}, i.e.~random rotation, scaling, and translation. 
We report additional information in Sec.~\ref{sec:implementation}.

\subsection{Network update}\label{sec:network_update}

We leverage the teacher-student learning scheme to facilitate the transfer of knowledge acquired during the course of the training with mixed domains.
We use the teacher network $\Phi_{\theta'}$ to produce target pseudo-labels $ \hat{\mathcal{Y}}^t_\textsf{U}$ for the student network $\Phi_{\theta}$, and train $\Phi_{\theta}$ to segment target point clouds by using the mixed point clouds $\mathcal{X}^{s \rightarrow t}$ and $\mathcal{X}^{t \rightarrow s}$ based on their mixed labels and pseudo-labels (Sec.~\ref{sec:compositional_mix}).

At each batch iteration, we update the student parameters $\Phi_\theta$ to minimize a total objective loss $\mathcal{L}_{tot}$ defined as

\begin{equation}\label{eq:total_loss}
    \mathcal{L}_{tot} = \mathcal{L}_{s \rightarrow t} + \mathcal{L}_{t \rightarrow s},
\end{equation}
where $\mathcal{L}_{s \rightarrow t}$ and $\mathcal{L}_{t \rightarrow s}$ are the $s \rightarrow t$ and $t \rightarrow s$ branch losses, respectively.
Given $\mathcal{X}^{s \rightarrow t}$ and  $\mathcal{Y}^{s \rightarrow t}$, we define the segmentation loss for the $s \rightarrow t$ branch as
\begin{equation}
    \mathcal{L}_{s \rightarrow t} = \mathcal{L}_{seg}(\Phi_{\theta}(\mathcal{X}^{s \rightarrow t}), \mathcal{Y}^{s \rightarrow t}),
\end{equation}
the objective of which is to minimize the segmentation error over $\mathcal{X}^{s \rightarrow t}$, thus learning to segment source patches in the target domain.
Similarly, given $\mathcal{X}^{t \rightarrow s}$ and  $\mathcal{Y}^{t \rightarrow s}$, we define the segmentation loss for the $t \rightarrow s$ branch as
\begin{equation}
    \mathcal{L}_{t \rightarrow s} = \mathcal{L}_{seg}(\Phi_{\theta}(\mathcal{X}^{t \rightarrow s}), \mathcal{Y}^{t \rightarrow s}),
\end{equation}
whose objective is to minimize the segmentation error over $\mathcal{X}^{t \rightarrow s}$ where target patches are composed with source data. 
We implement $\mathcal{L}_{seg}$ as the Dice segmentation loss~\cite{Jadon2020}, which we found effective for the segmentation of large-scale point clouds as it can cope with long-tail classes well.

Lastly, we update the teacher parameters $\theta'$ every $\gamma$ iterations following the exponential moving average (EMA)\cite{tarvainen2017mean} approach
\begin{equation}\label{eq:mean_teacher}
    \theta'_i = \beta\theta'_{i-1} + (1-\beta)\theta,
\end{equation}
where $i$ indicates the training iteration and $\beta$ is a smoothing coefficient hyperparamenter.



\section{Experiments}\label{sec:experimental}

We evaluate our method in both synthetic-to-real and real-to-real UDA and SSDA settings.
We use SynLiDAR~\cite{synlidar} as synthetic dataset, and SemanticKITTI~\cite{behley2019iccv, geiger2012we, geiger2013vision}, SemanticPOSS~\cite{pan2020semanticposs}, and nuScenes~\cite{nuscenes2019} as real-world datasets.
We compare \ourmethod with five state-of-the-art UDA methods: two general purpose adaptation methods (ADDA~\cite{tzeng2017adversarial}, Ent-Min~\cite{vu2019advent}), one image segmentation method (ST~\cite{zou2019confidence}), and two point cloud segmentation methods (PCT~\cite{synlidar}, ST-PCT~\cite{synlidar}).
Then, we compare \ourmethod with five state-of-the-art SSDA methods: three general purpose adaptation methods (MMD~\cite{tzeng2017adversarial}, MME~\cite{saito2019semi}, APE~\cite{kim2020attract}), and two point cloud segmentation methods (PCT~\cite{synlidar}, APE-PCT~\cite{synlidar}).
We refer to \ourmethod-UDA and \ourmethod-SSDA to indicate the version of CoSMix for UDA and SSDA, respectively, we use \ourmethod to refer to our method in general otherwise.
PCT, ST-PCT and APE-PCT are the only three state-of-the-art methods developed for 360$^\circ$ LiDAR point clouds and have only been applied for synthetic-to-real UDA and SSDA settings.
We re-implemented the comparison method and adapted to the same backbone network as that of CoSMix.
We refer to these methods as EntMin$^\star$~\cite{vu2019advent}, ST$^\star$~\cite{zou2019confidence}, MME$^\star$~\cite{saito2019semi}, MMD$^\star$~\cite{tzeng2017adversarial}, \textit{Source}$^\star$, \textit{Target}$^\star$ and \textit{Fine-tuned}$^\star$.
Moreover, we extended EntMin~\cite{vu2019advent} and ST~\cite{zou2019confidence} to the SSDA setting, and refer to them as EntMin-SSDA$^\star$ and ST-SSDA$^\star$.
For completeness, we also include the results of these methods as they are reported in~\cite{synlidar}.

\subsection{Datasets and metrics}\label{sec:dataset}

\noindent \textbf{SynLiDAR}~\cite{synlidar} is a large-scale synthetic dataset that is created with the Unreal Engine~\cite{Dosovitskiy17}.
It is composed of 198,396 annotated point clouds with 32 semantic classes. 
We use 19,840 point clouds for training and 1,976 point clouds for validation~\cite{synlidar}.
\noindent \textbf{SemanticPOSS}~\cite{pan2020semanticposs} is composed of 2,988 annotated real-world point cloud with 14 semantic classes.
We use the sequence $03$ for validation and the remaining sequences for training~\cite{pan2020semanticposs}.
For the SSDA settings, we follow~\cite{synlidar} and use the point cloud $172$ of sequence $02$ as the semi-supervised target set.
\noindent \textbf{SemanticKITTI}~\cite{behley2019iccv} is a large-scale segmentation dataset consisting of LiDAR acquisitions of the popular KITTI dataset~\cite{geiger2012we, geiger2013vision}.
It is composed of 43,552 annotated real-world point clouds with more than 19 semantic classes. 
We use sequence $08$ for validation and the remaining sequences for training~\cite{behley2019iccv}.
For the SSDA settings, we follow~\cite{synlidar} and use the point cloud $848$ from sequence $06$ and the point cloud $940$ from sequence $02$ as semi-supervised target set.
\noindent \textbf{nuScenes}~\cite{nuscenes2019} is a large-scale segmentation dataset. 
It is composed of real-world $850$ sequences ($700$ for training and $150$ for validation), for a total of $34,000$ annotated point clouds with 32 semantic classes. 
We use the official training and validation splits in all our experiments. 
For the SSDA settings, we follow the same selection protocol used in~\cite{synlidar} and use the point cloud with token n015-2018-07-24-11-13-19+0800\_\_LIDAR\_TOP\_\_1532402013197655 as semi-supervised target set.

We make source and target labels compatible across our datasets, i.e.~SynLiDAR $\rightarrow$ SemanticPOSS, SynLiDAR $\rightarrow$ SemanticKITTI and, SemanticKITTI $\rightarrow$ nuScenes.
In SynLiDAR $\rightarrow$ SemanticPOSS and SynLiDAR $\rightarrow$ SemanticKITTI, we follow \cite{synlidar} and map labels into $14$ segmentation classes and $19$ segmentation classes, respectively.
In SemanticKITTI $\rightarrow$ nuScenes we map source and target labels into $7$ common segmentation classes as in~\cite{saltori2022gipso}.

We evaluate the semantic segmentation performance before and after domain adaptation~\cite{synlidar} by using the Intersection over the Union (IoU)~\cite{rahman2016optimizing} for each segmentation class and report the per-class IoU.
We average the IoU over all the segmented classes and report the mean Intersection over the Union (mIoU).

\subsection{Implementation details}\label{sec:implementation}

We implemented \ourmethod in PyTorch and run our experiments on 4$\times$NVIDIA A100 (40GB SXM4).
We use MinkowskiNet as our point cloud segmentation network~\cite{choy20194d}, in particular we use MinkUNet32 as in~\cite{synlidar}. 
We pre-train our network on the source domain with Dice loss~\cite{Jadon2020} starting from randomly initialized weights.
In SSDA, we start from the pre-trained source model and finetune on both source and labeled target for two additional epochs. 
The finetuned model is used as pre-trained model in the semi-supervised settings.
In UDA, we initialize student and teacher networks with the parameters obtained after pre-training.
The pre-training and adaptation stage share the same hyperparameters.
In both the pre-training and adaptation steps, we use Stochastic Gradient Descent with a learning rate of $0.001$.

We set the value of $\alpha$ by examining the long-tailed classes present in the source domain during the adaptation process. 
Similarly, we set the parameter $\mu$ to the same value.
We assign the values of $\alpha$ and $\mu$ based on our prior experience, rather than optimizing these parameters through a systematic process.
In the target semantic selection function $g$, we establish the value of $\zeta$ based on a qualitative assessment of a few target frames, with the aim of producing spatially compact predictions. 
This approach yields approximately 80\% of pseudo-labeled points per scene.

On SynLiDAR $\rightarrow$ SemanticPOSS, we use a batch size of $12$ and perform adaptation for $10$ epochs. 
We set source and supervised target semantic selection ($f$) with $\alpha=0.5$ and $\mu=0.5$ while we set target semantic selection ($g$) with a confidence threshold $\zeta=0.85$.
On SynLiDAR $\rightarrow$ SemanticKITTI, we use a batch size of $16$, adapting for $3$ epochs.
During source and supervised target semantic selection ($f$) we set $\alpha=0.5$ and $\mu=0.5$ while in target semantic selection ($g$) we use a confidence threshold of $\zeta=0.90$. 
We use these last same hyperparameters also on SemanticKITTI $\rightarrow$ nuScenes, and SynLiDAR $\rightarrow$ nuScenes.

Our local augmentations $h$ and global augmentations $r$ are based on data augmentation strategies that are typical in the LiDAR segmentation literature~\cite{choy20194d}.
$h$ involves rigid rotation around the $z$-axis, scaling along all the axes and random point downsampling. 
We remove $xy$ rotation to produce co-planar and concentric mixed point clouds, and to preserve point ranges.
For the same reason, we remove rigid translations.
We bound rotations between $[-\pi/2, \pi/2]$ and scaling between $[0.95, 1.05]$, and perform random downsampling for $50\%$ of the patch points.
$r$ involves rigid rotation, translation and scaling along all the three axes. 
We set the parameters of $r$ the same as those used in~\cite{choy20194d}.
During the network update step (Sec.~\ref{sec:network_update}), we update the teacher parameters $\theta'_i$ with $\beta=0.99$.
On SynLiDAR $\rightarrow$ SemanticPOSS, we set $\gamma=1$ and do not perform parameter tuning.
On SynLiDAR $\rightarrow$ SemanticKITTI, we increase $\gamma$ to $\gamma=500$ to obtain a stable teacher behavior, i.e.~stable source performance, high average confidence of pseudo-labels, and $\sim80\%$ of pseudo-labeled points. 
We use these same hyperparameters also on SemanticKITTI $\rightarrow$ nuScenes, and SynLiDAR $\rightarrow$ nuScenes.

\subsection{Quantitative comparisons for UDA}\label{sec:results_uda}

\begin{table*}[t]
    \centering
    \caption{Unsupervised adaptation results on SynLiDAR $\rightarrow$ SemanticPOSS. We denote our reproduced baselines and results with $^\star$, e.g., \textit{Source}$^\star$. \textit{Source}$^\star$ and \textit{Target}$^\star$ correspond to the model trained on the source synthetic dataset (lower bound) and on the target real dataset (upper bound), respectively. Results are reported in terms of mean Intersection over the Union (mIoU).}
    \label{tab:adaptation_poss}
    \tabcolsep 4pt
    \resizebox{.8\textwidth}{!}{%
    \begin{tabular}{l|ccccccccccccc|c}
        \toprule
        \textbf{Model} & \textbf{pers.} & \textbf{rider} & \textbf{car} & \textbf{trunk} & \textbf{plants} & \textbf{traf.} & \textbf{pole} & \textbf{garb.} & \textbf{buil.} & \textbf{cone.} & \textbf{fence} & \textbf{bike} & \textbf{grou.} & \textbf{mIoU} \\
        \midrule
        
        \textit{Source} & 3.7 & 25.1 & 12.0 & 10.8 & 53.4 & 0.0 & 19.4 & 12.9 & 49.1 & 3.1 & 20.3 & 0.0 & 59.6 & 20.7 \\
        \textit{Source}$^\star$ & 21.7 & 20.1 & 9.7 & 3.4 & 56.8 & 4.8 & 24.1 & 6.1 & 39.9 & 0.3 & 15.3 & 5.3 & 73.4 & 21.6\\
        \textit{Target}$^\star$ & 61.8 & 54.7 & 33.0 & 19.3 & 73.9 & 26.7 & 30.9 & 11.0 & 71.3 & 32.5 & 44.6 & 43.2 & 78.5 & 44.7 \\
        \midrule
        ADDA~\cite{tzeng2017adversarial} & 27.5 & 35.1 & 18.8 & 12.4 & 53.4 & 2.8 & 27.0 & 12.2 & 64.7 & 1.3 & 6.3 & 6.8 & 55.3 & 24.9 \\
        Ent-Min~\cite{vu2019advent} & 24.2 & 32.2 & 21.4 & 18.9 & 61.0 & 2.5 & 36.3 & 8.3 & 56.7 & 3.1 & 5.3 & 4.8 & 57.1 & 25.5 \\
        Ent-Min$^\star$~\cite{vu2019advent} & 24.8 & 28.0 & 13.4 & 4.1 & 59.6 & 2.0 & 23.3 & 5.8 & 47.0 & 0.0 & 16.1 & 5.8 & 71.6 & 23.2\\
        ST~\cite{zou2019confidence} & 23.5 & 31.8 & 22.0 & 18.9 & 63.2 & 1.9 & \textbf{41.6} & 13.5 & 58.2 & 1.0 & 9.1 & 6.8 & 60.3 & 27.1\\
        ST$^\star$~\cite{zou2019confidence} & 47.7 & 42.6 & 24.4 & 13.8 & 62.5 & 3.3 & 36.1 & 23.5 & 50.9 & 18.8 & 14.6 & 4.0 & 68.9 & 31.6\\
        PCT~\cite{synlidar} & 13.0 & 35.4 & 13.7 & 10.2 & 53.1 & 1.4 & 23.8 & 12.7 & 52.9 & 0.8 & 13.7 & 1.1 & 66.2 & 22.9\\
        ST-PCT~\cite{synlidar} & 28.9 & 34.8 & 27.8 & 18.6 & 63.7 & 4.9 & 41.0 & 16.6 & 64.1 & 1.6 & 12.1 & 6.6 & 63.9 & 29.6\\
        \midrule
        \ourmethod-UDA & \textbf{55.8} & \textbf{51.4} & \textbf{36.2} & \textbf{23.5} & \textbf{71.3} & \textbf{22.5} & 34.2 & \textbf{28.9} & \textbf{66.2} & \textbf{20.4} & \textbf{24.9} & \textbf{10.6} & \textbf{78.7} & \textbf{40.4}\\
        \bottomrule
    \end{tabular}
    }
\end{table*}

\begin{table*}[t]
    \centering
    \caption{Unsupervised adaptation results on SynLiDAR $\rightarrow$ SemanticKITTI. We denote our reproduced baselines and results with $^\star$, e.g., \textit{Source}$^\star$. \textit{Source}$^\star$ and \textit{Target}$^\star$ correspond to the model trained on the source synthetic dataset (lower bound) and on the target real dataset (upper bound), respectively. Results are reported in terms of mean Intersection over the Union (mIoU).}
    \label{tab:adaptation_kitti}
    \tabcolsep 5pt
    \resizebox{\textwidth}{!}{%
    \begin{tabular}{l|ccccccccccccccccccc|c}
        \toprule
        \textbf{Model} & \rotatebox{90}{\textbf{car}} & \rotatebox{90}{\textbf{bi.cle}} & \rotatebox{90}{\textbf{mt.cle}} & \rotatebox{90}{\textbf{truck}} & \rotatebox{90}{\textbf{oth-v.}} & \rotatebox{90}{\textbf{pers.}} & \rotatebox{90}{\textbf{b.clst}} & \rotatebox{90}{\textbf{m.clst}} & \rotatebox{90}{\textbf{road}} & \rotatebox{90}{\textbf{park.}} & \rotatebox{90}{\textbf{sidew.}} & \rotatebox{90}{\textbf{oth-g.}} & \rotatebox{90}{\textbf{build.}} & \rotatebox{90}{\textbf{fence}} & \rotatebox{90}{\textbf{veget.}} & \rotatebox{90}{\textbf{trunk}} & \rotatebox{90}{\textbf{terra.}} & \rotatebox{90}{\textbf{pole}} & \rotatebox{90}{\textbf{traff.}} & \textbf{mIoU} \\
        \midrule
        
        \textit{Source} & 42.0 & 5.0 & 4.8 & 0.4 & 2.5 & 12.4 & 43.3 & 1.8 & 48.7 & 4.5 & 31.0 & 0.0 & 18.6 & 11.5 & 60.2 & 30.0 & 48.3 & 19.3 & 3.0 & 20.4 \\
        \textit{Source}$^\star$ & 60.7 & 1.9 & 22.0 & 10.3 & 8.0 & 16.7 & 11.3 & 20.3 & 70.4 & 6.4 & 40.4 & 0.0 & 25.6 & 8.6 & 59.5 & 18.4 & 29.1 & 29.0 & 13.9 & 23.8\\
        \textit{Target}$^\star$ & 90.0 & 6.3 & 20.3 & 63.0 & 18.1 & 31.1 & 39.6 & 5.8 & 90.9 & 29.0 & 74.7 & 4.0 & 85.4 & 23.3 & 83.9 & 46.2 & 62.2 & 40.7 & 20.6 & 44.0\\
        \midrule

        ADDA~\cite{tzeng2017adversarial} & 52.5 & 4.5 & 11.9 & 0.3 & 3.9 & 9.4 & 27.9 & 0.5 & 52.8 & 4.9 & 27.4 & 0.0 & 61.0 & 17.0 & 57.4 & 34.5 & 42.9 & 23.2 & 4.5 & 23.0\\
        Ent-Min~\cite{vu2019advent} & 58.3 & 5.1 & 14.3 & 0.3 & 1.8 & 14.3 & \textbf{44.5} & 0.5 & 50.4 & 4.3 & 34.8 & 0.0 & 48.3 & 19.7 & 67.5 & 34.8 & \textbf{52.0} & 33.0 & 6.1 & 25.8 \\
        Ent-Min$^\star$~\cite{vu2019advent} & 63.8 & \textbf{8.5} & 23.0 & 15.9 & 5.0 & 17.2 & 33.3 & 22.8 & 61.6 & 3.1 & 34.4 & \textbf{0.2} & 52.2 & 6.2 & 63.3 & 16.9 & 19.9 & 27.5 & 9.4 & 25.5\\
        ST~\cite{zou2019confidence} & 62.0 & 5.0 & 12.4 & 1.3 & 9.2 & 16.7 & 44.2 & 0.4 & 53.0 & 2.5 & 28.4 & 0.0 & 57.1 & 18.7 & 69.8 & \textbf{35.0} & 48.7 & 32.5 & 6.9 & 26.5 \\
        ST$^\star$~\cite{zou2019confidence} & 69.7 & 6.4 & 18.3 & 4.4 & 5.8 & 14.8 & 23.3 & 20.2 & 54.2 & 5.3 & 34.1 & 0.1 & 44.3 & 5.1 & 63.5 & 16.8 & 26.9 & 30.6 & 12.2 & 24.0\\
        PCT~\cite{synlidar} & 53.4 & 5.4 & 7.4 & 0.8 & 10.9 & 12.0 & 43.2 & 0.3 & 50.8 & 3.7 & 29.4 & 0.0 & 48.0 & 10.4 & 68.2 & 33.1 & 40.0 & 29.5 & 6.9 & 23.9 \\
        ST-PCT~\cite{synlidar} & 70.8 & 7.3 & 13.1 & 1.9 & 8.4 & 12.6 & 44.0 & 0.6 & 56.4 & 4.5 & 31.8 & 0.0 & \textbf{66.7} & \textbf{23.7} & \textbf{73.3} & 34.6 & 48.4 & \textbf{39.4} & 11.7 & 28.9 \\
        \midrule
        \ourmethod-UDA & \textbf{75.1} & 6.8 & \textbf{29.4} & \textbf{27.1} & \textbf{11.1} & \textbf{22.1} & 25.0 & \textbf{24.7} & \textbf{79.3} & \textbf{14.9} & \textbf{46.7} & 0.1 & 53.4 & 13.0 & 67.7 & 31.4 & 32.1 & 37.9 & \textbf{13.4} & \textbf{32.2}\\
        \bottomrule

    \end{tabular}
    }
\end{table*}

\begin{table}[t]
    \centering
    \caption{Unsupervised adaptation results on SemanticKITTI $\rightarrow$ nuScenes. We denote our reproduced baselines and results with $^\star$, e.g., \textit{Source}$^\star$. \textit{Source}$^\star$ and \textit{Target}$^\star$ correspond to the model trained on the source real dataset (lower bound) and on the target real dataset (upper bound), respectively. Results are reported in terms of mean Intersection over the Union (mIoU).
    }
    \label{tab:adaptation_uda_real2real}
    \tabcolsep 4pt
    \resizebox{1\columnwidth}{!}{%
    \begin{tabular}{l|ccccccc|c}
        \toprule
        \textbf{Model} & \rotatebox{90}{\textbf{car}} & \rotatebox{90}{\textbf{pers.}} & \rotatebox{90}{\textbf{road}} & \rotatebox{90}{\textbf{side.}} & \rotatebox{90}{\textbf{terr.}} & \rotatebox{90}{\textbf{manm.}} & \rotatebox{90}{\textbf{vege.}} & \textbf{mIoU} \\
        \midrule
        \textit{Source}$^\star$ & 29.4 & 15.6 & 73.2 & 29.1 & 14.7 & 58.5 & 59.9 & 40.1\\
        \textit{Target}$^\star$ & 35.8 & 43.2 & 93.6 & 62.1 & 49.0 & 76.4 & 73.9 & 62.0\\

        \midrule
        EntMin$^\star$\cite{vu2019advent} & \textbf{33.3} & 12.6 & \textbf{78.3} & \textbf{35.7} & 18.4 & 63.1 & 62.4 & 43.4\\
        ST$^\star$\cite{zou2019confidence} & 30.6 & 20.6 & 79.1 & 34.4 & 18.9 & 62.4 & 59.3 & 43.6\\
        \midrule
        \ourmethod-UDA & 32.1 & \textbf{26.3} & 78.1 & 35.1 & \textbf{20.2} & \textbf{66.4} & \textbf{65.2} & \textbf{46.2}\\
        \bottomrule
    \end{tabular}
    }
\end{table}

\noindent\textbf{Synthetic-to-real.}
Tabs.~\ref{tab:adaptation_poss}\&\ref{tab:adaptation_kitti} report the results in the UDA settings on SynLiDAR $\rightarrow$ SemanticPOSS, and on SynLiDAR $\rightarrow$ SemanticKITTI, respectively.
The \textit{Source}$^\star$ model is the lower bound of each scenario with $21.6$ mIoU on SynLiDAR $\rightarrow$ SemanticPOSS and $23.8$ mIoU on SynLiDAR $\rightarrow$ SemanticKITTI. 
The \textit{Target}$^\star$ model is the upper bound of each scenario with $44.7$ mIoU on SynLiDAR $\rightarrow$ SemanticPOSS and $44.0$ mIoU on SynLiDAR $\rightarrow$ SemanticKITTI.
Note that \textit{Source}$^\star$ models always outperform \textit{Source}. 
This may be due to a better parameter choice that leads an improved generalization ability.
In SynLiDAR $\rightarrow$ SemanticPOSS (Tab.~\ref{tab:adaptation_poss}), \ourmethod-UDA outperforms the other methods on all the classes, except on \textit{pole} where ST achieves a better result.
On average, we achieve $40.4$ mIoU, surpassing ST-PCT by $+10.8$ mIoU and improving over the \textit{Source}$^\star$ of $+18.8$ mIoU. 
\ourmethod-UDA improves also on difficult classes as \textit{person}, \textit{traffic-sign}, \textit{cone}, and \textit{bike}, whose performance are rather low before domain adaptation.
ST$^\star$ and EntMin$^\star$ improve over \textit{Source}$^\star$. ST$^\star$ improves over ST while EntMin$^\star$ achieves lower performance.
Tab.~\ref{tab:adaptation_kitti} reports the results of SynLiDAR $\rightarrow$ SemanticKITTI. SemanticKITTI is challenging as the validation sequence includes a wide range of different scenarios with a large number of semantic classes.
\ourmethod-UDA improves all the classes when compared to Source$^\star$, except for \textit{traffic-cone}. 
We believe this is due to the noise introduced by the pseudo labels on these classes and in related classes such as $road$.
\ourmethod-UDA improves on $10$ out of $19$ classes, with a large margin in the classes \textit{car}, \textit{motorcycle}, \textit{truck}, \textit{person}, \textit{road}, \textit{parking} and \textit{sidewalk}.
On average, we achieve state-of-the-art performance with a $32.2$ mIoU, outperforming ST-PCT by $+3.3$ mIoU and improving over Source$^\star$ of about $+8.4$ mIoU.

\vspace{1mm}
\noindent\textbf{Real-to-real.}
Tab.~\ref{tab:adaptation_uda_real2real} reports the results on SemanticKITTI $\rightarrow$ nuScenes in the UDA setting.
SemanticKITTI $\rightarrow$ nuScenes is a more challenging direction as source and target sensors are different, and nuScenes has rather sparse point clouds.
\textit{Source}$^\star$ and \textit{Target}$^\star$ models achieve $40.1$ mIoU and $62.0$ mIoU, respectively. 
\ourmethod-UDA outperforms the compared methods on $4$ out of $7$ classes, with the largest margin on the class \textit{person}.
On average, EntMin$^\star$ and ST$^\star$ achieve $43.4$ mIoU and $43.6$ mIoU, showing a limited improvement over \textit{Source}$^\star$.
\ourmethod-UDA achieves the best results of $46.2$ mIoU, outperforming all the compared methods.

\subsection{Quantitative comparison for SSDA}\label{sec:results_ssda}

\noindent\textbf{Synthetic-to-real.}
Tabs.~\ref{tab:adaptation_ssda_poss}\&\ref{tab:adaptation_ssda_kitti} report the results in the SSDA settings on SynLiDAR $\rightarrow$ SemanticPOSS, and on SynLiDAR $\rightarrow$ SemanticKITTI, respectively.
\textit{Source}$^\star$ and \textit{Target}$^\star$ models are the lower and upper bound of the UDA settings.
The \textit{Fine-tuned}$^\star$ model is obtained by fine-tuning \textit{Source}$^\star$ with the semi-supervised target samples. 
It shows the highest possible bound without any adaptation approach. 
\textit{Fine-tuned}$^\star$ always outperforms \textit{Fine-tuned} from~\cite{synlidar}. 
Similarly, the discrepancy between MMD$^\star$ and MME$^\star$, and the results reported in~\cite{synlidar} may due to a different parameter choice.
In SynLiDAR $\rightarrow$ SemanticPOSS (Tab.~\ref{tab:adaptation_ssda_poss}), \ourmethod-SSDA outperforms all the comparison methods on all the classes, except on \textit{plants}, \textit{fence} and \textit{bike} where MME and MME$^\star$ achieve better results.
On average, we reach $41.0$ mIoU, outperforming APE-PCT by $+9.8$ mIoU and improving over \textit{Source}$^\star$ by $+19.4$ and over \textit{Fine-tuned}$^\star$ by $+15.5$. 
Compared to \ourmethod-UDA, \ourmethod-SSDA achieves a $+0.6$ mIoU, getting closer to the \textit{Target} upper bound.
In SynLiDAR $\rightarrow$ SemanticKITTI (Tab.~\ref{tab:adaptation_ssda_kitti}), \ourmethod-SSDA brings a significant improvement on $12$ out $19$, especially on \textit{car}, \textit{truck}, \textit{motorcyclist}, \textit{road}, \textit{parking}, and \textit{pole}. 
On average, \ourmethod-SSDA achieves $34.3$ mIoU, outperforming the best baseline APE-PCT by $+7.3$ mIoU and improving over \textit{Source}$^\star$ of $+10.5$ mIoU and over \textit{Fine-tuned}$^\star$ of $+9.8$ mIoU. 
Compared to our UDA pipeline, \ourmethod-SSDA improves of $+2.1$ mIoU, showing that the additional target supervision is beneficial for further reducing the domain gap.

\begin{table*}[t]
    \centering
    \caption{Semi-supervised adaptation results on SynLiDAR $\rightarrow$ SemanticPOSS. We denote our reproduced baselines and results with $^\star$, e.g., \textit{Source}$^\star$. \textit{Source}$^\star$ and \textit{Target}$^\star$ correspond to the model trained on the source synthetic dataset (lower bound) and on the target real dataset (upper bound), respectively. Results are reported in terms of mean Intersection over the Union (mIoU).
    }
    \label{tab:adaptation_ssda_poss}
    \tabcolsep 4pt
    \resizebox{.8\textwidth}{!}{%
    \begin{tabular}{l|ccccccccccccc|c}
        \toprule
        \textbf{Model} & \textbf{pers.} & \textbf{rider} & \textbf{car} & \textbf{trunk} & \textbf{plants} & \textbf{traf.} & \textbf{pole} & \textbf{garb.} & \textbf{buil.} & \textbf{cone.} & \textbf{fence} & \textbf{bike} & \textbf{grou.} & \textbf{mIoU} \\
        \midrule
        \textit{Source} & 3.7 & 25.1 & 12.0 & 10.8 & 53.4 & 0.0 & 19.4 & 12.9 & 49.1 & 3.1 & 20.3 & 0.0 & 59.6 & 20.7\\
        \textit{Source}$^\star$ & 21.7 & 20.1 & 9.7 & 3.4 & 56.8 & 4.8 & 24.1 & 6.1 & 39.9 & 0.3 & 15.3 & 5.3 & 73.4 & 21.6\\
        \textit{Fine-tuned} & 25.2 & 36.1 & 18.2 & 12.8 & 58.6 & 1.7 & 30.5 & 5.6 & 25.7 & 3.0 & 12.0 & 10.6 & 75.6 & 24.3\\
        \textit{Fine-tuned}$^\star$ & 25.2 & 27.2 & 21.6 & 9.6 & 60.4 & 0.7 & 16.2 & 10.8 & 44.5 & 12.0 & 24.1 & 2.5 & 76.7 & 25.5\\
        \textit{Target}$^\star$ & 61.8 & 54.7 & 33.0 & 19.3 & 73.9 & 26.7 & 30.9 & 11.0 & 71.3 & 32.5 & 44.6 & 43.2 & 78.5 & 44.7 \\
        
        \midrule
        MMD~\cite{tzeng2017adversarial} & 25.5 & 35.7 & 28.9 & 6.7 & 64.3 & 1.7 & 23.2 & 5.6 & 53.3 & 3.3 & 30.2 & 13.9 & 70.4 & 27.9 \\
        MMD$^\star$\cite{tzeng2017adversarial} & 28.1 & 12.2 & 18.8 & 11.4 & 71.5 & 10.0 & 14.7 & 0.0 & 64.6 & 0.0 & 28.1 & 25.1 & 78.6 & 27.9\\
        MME~\cite{saito2019semi} & 33.2 & 40.2 & 25.0 & 11.0 & 61.9 & 0.4 & 31.2 & 7.3 & 56.1 & 5.7 & \textbf{37.1} & 6.7 & 71.2 & 29.8\\
        MME$^\star$\cite{saito2019semi} & 35.8 & 16.1 & 21.4 & 7.9 & \textbf{73.7} & 7.9 & 24.2 & 1.5 & 67.6 & 0.0 & 32.8 & \textbf{32.0} & 77.0 & 30.6\\
        APE~\cite{kim2020attract} & 34.3 & 40.1 & 21.5 & 16.3 & 62.6 & 0.9 & 31.1 & 2.3 & 55.9 & 13.3 & 34.3 & 9.6 & 71.6 & 30.3 \\
        EntMin-SSDA$^\star$ & 24.7 & 9.4 & 16.5 & 10.7 & 69.9 & 6.7 & 11.7 & 0.0 & 62.6 & 0.0 & 25.2 & 22.5 & 78.9 & 26.1\\
        ST-SSDA$^\star$ & 40.1 & 24.3 & 22.5 & 7.9 & 70.2 & 13.4 & 21.7 & 1.4 & 66.9 & 0.1 & 34.7 & \textbf{32.0} & 78.1 & 31.8\\
        PCT~\cite{synlidar} & 25.8 & 36.8 & 27.8 & 11.3 & 62.2 & 1.9 & 31.2 & 5.2 & 58.7 & 2.6 & 34.3 & 8.5 & 68.7 & 28.8\\
        APE-PCT~\cite{synlidar} & 34.7 & 36.3 & 27.2 & 15.8 & 62.9 & 0.8 & 31.6 & 8.7 & 62.3 & 9.8 & 35.1 & 9.3 & 70.9 & 31.2\\
        \midrule
        \ourmethod-SSDA & \textbf{54.9} & \textbf{50.6} & \textbf{33.4} & \textbf{22.5} & 73.0 & \textbf{13.6} & \textbf{38.4} & \textbf{26.4} & \textbf{68.5} & \textbf{16.2} & 29.6 & 27.4 & \textbf{79.0} & \textbf{41.0}\\
        \bottomrule
    \end{tabular}
    }
\end{table*}

\begin{table*}[t]
    \centering
    \caption{Semi-supervised adaptation results on SynLiDAR $\rightarrow$ SemanticKITTI. We denote our reproduced baselines and results with $^\star$, e.g., \textit{Source}$^\star$. \textit{Source}$^\star$ and \textit{Target}$^\star$ correspond to the model trained on the source synthetic dataset (lower bound) and on the target real dataset (upper bound), respectively. Results are reported in terms of mean Intersection over the Union (mIoU).
    }
    \label{tab:adaptation_ssda_kitti}
    \tabcolsep 5pt
    \resizebox{\textwidth}{!}{%
    \begin{tabular}{l|ccccccccccccccccccc|c}
        \toprule
        \textbf{Model} & \rotatebox{90}{\textbf{car}} & \rotatebox{90}{\textbf{bi.cle}} & \rotatebox{90}{\textbf{mt.cle}} & \rotatebox{90}{\textbf{truck}} & \rotatebox{90}{\textbf{oth-v.}} & \rotatebox{90}{\textbf{pers.}} & \rotatebox{90}{\textbf{b.clst}} & \rotatebox{90}{\textbf{m.clst}} & \rotatebox{90}{\textbf{road}} & \rotatebox{90}{\textbf{park.}} & \rotatebox{90}{\textbf{sidew.}} & \rotatebox{90}{\textbf{oth-g.}} & \rotatebox{90}{\textbf{build.}} & \rotatebox{90}{\textbf{fence}} & \rotatebox{90}{\textbf{veget.}} & \rotatebox{90}{\textbf{trunk}} & \rotatebox{90}{\textbf{terra.}} & \rotatebox{90}{\textbf{pole}} & \rotatebox{90}{\textbf{traff.}} & \textbf{mIoU} \\
        \midrule
        \textit{Source} &  42.0 & 5.0 & 4.8 & 0.4 & 2.5 & 12.4 & 43.3 & 1.8 & 48.7 & 4.5 & 31.0 & 0.0 & 18.6 & 11.5 & 60.2 & 30.0 & \textbf{48.3} & 19.3 & 3.0 & 20.4\\
        \textit{Source}$^\star$ &  60.7 & 1.9 & 22.0 & 10.3 & 8.0 & 16.7 & 11.3 & 20.3 & 70.4 & 6.4 & 40.4 & 0.0 & 25.6 & 8.6 & 59.5 & 18.4 & 29.1 & 29.0 & 13.9 & 23.8\\
        \textit{Fine-tuned} & 56.2 & 3.0 & 15.1 & 1.0 & 5.0 & 20.2 & 42.1 & 2.8 & 52.1 & 0.7 & 19.8 & 0.0 & 41.3 & 5.8 & 62.1 & 34.0 & 42.0 & 24.6 & 1.4 & 22.6\\
        \textit{Fine-tuned}$^\star$ & 61.8 & 2.8 & 21.7 & 10.8 & 4.2 & 14.5 & 18.0 & 15.9 & 65.6 & 6.4 & 40.2 & 0.0 & 34.2 & 7.0 & 60.6 & 22.1 & 39.8 & 32.2 & 8.4 & 24.5\\
        \textit{Target}$^\star$ & 90.0 & 6.3 & 20.3 & 63.0 & 18.1 & 31.1 & 39.6 & 5.8 & 90.9 & 29.0 & 74.7 & 4.0 & 85.4 & 23.3 & 83.9 & 46.2 & 62.2 & 40.7 & 20.6 & 44.0\\
        
        \midrule
        MMD~\cite{tzeng2017adversarial} & 56.4 & 3.3 & 13.3 & 1.5 & 6.1 & 21.4 & 34.6 & 1.6 & 54.3 & 0.4 & 21.4 & 0.0 & 50.2 & 5.8 & 61.2 & \textbf{37.0} & 44.9 & 31.6 & 2.2 & 23.5\\
        MMD$^\star$\cite{tzeng2017adversarial} & 46.5 & 3.2 & 6.3 & 12.1 & 3.3 & 8.8 & 21.7 & 13.4 & 47.2 & 4.6 & 29.9 & 0.0 & 50.6 & 5.9 & 62.2 & 16.2 & 23.3 & 19.4 & 4.7 & 20.0\\
        MME~\cite{saito2019semi} & 51.0 & 5.6 & 13.1 & 1.3 & 7.3 & 15.1 & 54.4 & 4.4 & 43.1 & 0.2 & 28.3 & 0.0 & \textbf{60.7} & 13.3 & 66.1 & 30.1 & 39.9 & 24.8 & 6.6 & 24.5\\
        MME$^\star$\cite{saito2019semi} & 28.7 & 0.2 & 1.0 & 1.8 & 0.9 & 2.0 & 1.6 & 3.5 & 53.6 & 1.8 & 31.1 & 0.0 & 40.6 & 7.2 & 57.6 & 12.1 & 26.7 & 14.4 & 0.1 & 15.0\\
        APE~\cite{kim2020attract} & 58.6 & 6.2 & 16.6 & 3.1 & 11.3 & 14.2 & 35.8 & 3.7 & 61.5 & 1.7 & 30.3 & 0.0 & 54.7 & \textbf{15.4} & 64.6 & 20.0 & \textbf{45.5} & 23.9 & 9.1 & 25.1\\
        EntMin-SSDA$^\star$ & 52.0 & 2.6 & 7.8 & 10.3 & 3.5 & 8.4 & 20.9 & 13.2 & 42.1 & 3.5 & 31.2 & 0.0 & 44.1 & 5.8 & 62.5 & 15.2 & 22.7 & 18.2 & 5.8 & 19.5\\
        ST-SSDA$^\star$ & 60.0 & 2.8 & 10.6 & 14.6 & 5.0 & 10.4 & 19.2 & 20.8 & 63.9 & 4.5 & 35.7 & 0.1 & 43.4 & 7.1 & 62.2 & 13.3 & 26.9 & 24.8 & 10.4 & 22.9\\
        PCT~\cite{synlidar} & 56.0 & 7.0 & 17.1 & 2.8 & 9.9 & 23.7 & 43.7 & 5.6 & 55.3 & 0.8 & 22.9 & 0.0 & 50.1 & 8.4 & 65.3 & 23.1 & 43.5 & 28.8 & 7.5 & 24.8\\
        APE-PCT~\cite{synlidar} & 58.1 & 7.3 & 17.8 & 2.6 & \textbf{13.9} & \textbf{24.7} & \textbf{46.5} & 5.1 & 60.5 & 1.9 & 31.3 & 0.0 & 56.8 & 14.6 & 67.9 & 23.7 & 44.3 & 26.1 & 9.3 & 27.0\\
        \midrule
        \ourmethod-SSDA & \textbf{76.9} & \textbf{10.4} & \textbf{27.1} & \textbf{23.1} & 13.4 & 24.0 & 21.7 & \textbf{27.9} & \textbf{75.8} & \textbf{17.9} & \textbf{49.7} & \textbf{0.1} & 60.3 & 
        14.7 & \textbf{69.8} & 36.8 & 40.9 & \textbf{45.6} & \textbf{16.2} & \textbf{34.3}\\

        \bottomrule

    \end{tabular}
    }
\end{table*}

\begin{table}[t]
    \centering
    \caption{Semi-supervised adaptation results on SemanticKITTI $\rightarrow$ nuScenes. We denote our reproduced baselines and results with $^\star$, e.g., \textit{Source}$^\star$. \textit{Source}$^\star$ and \textit{Target}$^\star$ correspond to the model trained on the source real dataset (lower bound) and on the target real dataset (upper bound), respectively. Results are reported in terms of mean Intersection over the Union (mIoU).}
    \label{tab:adaptation_ssda_real2real}
    \tabcolsep 4pt
    \resizebox{1\columnwidth}{!}{%
    \begin{tabular}{l|ccccccc|c}
        \toprule
        \textbf{Model} & \rotatebox{90}{\textbf{car}} & \rotatebox{90}{\textbf{pers.}} & \rotatebox{90}{\textbf{road}} & \rotatebox{90}{\textbf{side.}} & \rotatebox{90}{\textbf{terr.}} & \rotatebox{90}{\textbf{manm.}} & \rotatebox{90}{\textbf{vege.}} & \textbf{mIoU} \\
        \midrule
        \textit{Source}$^\star$ & 29.4 & 15.6 & 73.2 & 29.1 & 14.7 & 58.5 & 59.9 & 40.1\\
        \textit{Fine-tuned}$^\star$ & 47.0 & 22.9 & 75.1 & 28.6 & 15.3 & 61.8 & 53.7 & 43.5\\
        \textit{Target}$^\star$ & 35.8 & 43.2 & 93.6 & 62.1 & 49.0 & 76.4 & 73.9 & 62.0\\
        \midrule
        MMD$^\star$~\cite{tzeng2017adversarial} & 38.3 & 14.1 & 83.2 & 32.4 & 33.9 & 63.7 & 63.2 & 47.0\\
        MME$^\star$~\cite{saito2019semi} & 41.0 & 9.5 & \textbf{83.9} & 31.7 & 32.9 & 63.3 & 57.8 & 45.7\\
        EntMin-SSDA$^\star$ & 31.8 & 13.6 & 81.8 & \textbf{35.5} & 30.7 & 66.2 & 65.7 & 46.5\\
        ST-SSDA$^\star$ & 44.9 & 13.9 & 71.9 & 22.6 & \textbf{34.1} & \textbf{68.0} & \textbf{67.7} & 46.2\\
        \midrule
        \ourmethod-SSDA & \textbf{45.3} & \textbf{26.9} & 80.1 & 34.5 & 20.8 & \textbf{68.0} & 67.0 & \textbf{48.9}\\
        \bottomrule
    \end{tabular}
    }
\end{table}

\vspace{1mm}
\noindent\textbf{Real-to-real.}
Tab.~\ref{tab:adaptation_ssda_real2real} reports the results on SemanticKITTI $\rightarrow$ nuScenes in the SSDA settings.
\textit{Source}$^\star$ and \textit{Target}$^\star$ models achieve $40.1$ mIoU and $62.0$ mIoU, respectively. 
The \textit{Fine-tuned}$^\star$ model improves over \textit{Source}$^\star$ and achieves $43.5$ mIoU.
\ourmethod-SSDA achieves the best results on $3$ out of $7$ classes, with the largest margin on the class \textit{pedestrian}.
On average, MMD$^\star$ is the best performing method among the comparison methods with $47.0$ mIoU. 
\ourmethod-SSDA achieves $48.9$ mIoU, outperforms all the comparison methods and further improving over \ourmethod-UDA.

\subsection{Domain adaptation between different sensors}\label{sec:ablation_synth2nusc}

\begin{table*}[t]
    \centering
    \caption{Adaptation results on SynLiDAR$\rightarrow$nuScenes. We denote our reproduced baselines and results with $^\star$, e.g., \textit{Source}$^\star$. \textit{Source}$^\star$ and \textit{Target}$^\star$ correspond to the model trained on the source synthetic dataset (lower bound) and on the target real dataset (upper bound), respectively. Results are reported in terms of mean Intersection over the Union (mIoU).}
    \label{tab:adaptation_synth2nusc}
    \tabcolsep 4pt
    \resizebox{0.9\textwidth}{!}{%
    \begin{tabular}{l|ccccccccccc|c}
        \toprule
        \textbf{Model} & \textbf{car} &	\textbf{bicycle} & \textbf{motorcycle} & \textbf{truck} & \textbf{bus} & \textbf{pedestrian} & \textbf{road} & \textbf{sidewalk} & \textbf{building} &	\textbf{vegetation} & \textbf{terrain} & \textbf{mIoU} \\
        \midrule
        \textit{Source}$^\star$ & 23.7 & 2.8 & 10.3 & 15.8 & 4.9 & 20.8 & 63.6 & 18.0 & 47.7 & 50.0 & 3.3 & 23.7\\
        \textit{Fine-tuned}$^\star$ & 27.9 & 3.4 & 14.2 & 15.9 & 5.0 & 22.1 & 66.5 & 18.6 & 58.8 & 54.5 & 3.9 & 26.4\\
        \textit{Target}$^\star$ & 38.4 & 12.3 & 22.2 & 33.4 & 37.0 & 40.1 & 91.6 & 56.7 & 75.3 & 71.6 & 46.0 & 47.7\\
        \midrule
        \ourmethod-UDA & 27.8 & \textbf{4.8} & \textbf{11.9} & \textbf{16.1} & 6.4 & \textbf{24.2} & 67.1 & 19.6 & 60.3 & 59.2 & 2.7 & 27.3\\
        \ourmethod-SSDA & \textbf{28.8} & 4.6 & 7.8 & 11.1 & \textbf{10.1} & 22.7 & \textbf{72.8} & \textbf{21.6} & \textbf{60.4} & \textbf{60.5} & \textbf{3.2} & \textbf{27.6}\\
        \bottomrule
    \end{tabular}
    }
\end{table*}

We study the domain adaptation performance of \ourmethod when the source point cloud is synthetically generated with a certain sensor and the target point cloud is captured in the real world with a different sensor.
We use SynLiDAR as the source domain and nuScenes as the target domain. 
The semantic classes are mapped into the common $11$ segmentation classes: \textit{car}, \textit{bicycle}, \textit{motorcycle}, \textit{truck}, \textit{bus}, \textit{pedestrian}, \textit{road}, \textit{sidewalk}, \textit{building}, \textit{vegetation} and, \textit{terrain}.
This case exhibits a rather strong domain shift, as the SynLiDAR point clouds are dense and nearly noise-free, while the nuScenes point clouds are sparser and noisier.
We follow the same implementation details of SemanticKITTI $\rightarrow$ nuScenes, except we change the target domain.
Tab.~\ref{tab:adaptation_synth2nusc} reports the domain adaptation results on SynLiDAR $\rightarrow$ nuScenes in both the UDA and SSDA settings.
\textit{Source}$^\star$ and \textit{Target}$^\star$ models achieve $23.7$ mIoU and $47.7$ mIoU, respectively. 
\textit{Fine-tuned}$^\star$ improves the performance to $26.4$ mIoU.
\ourmethod-UDA improves over \textit{Source}$^\star$ by achieving $27.3$ mIoU. 
Despite the lack of target supervision, we also outperform the \textit{Fine-tuned}$^\star$ baseline. 
\ourmethod-SSDA further improves the results by achieving $27.6$ mIoU when we introduce limited target supervision. 
We observed a lower improvement of \ourmethod compared to the other adaptation directions, which we attribute to the different simulated sensor and the large density difference between SynLiDAR and nuScenes scans.

\subsection{Qualitative results}\label{sec:qualitative}

\begin{figure*}[t]
\centering
    \setlength\tabcolsep{1.pt}
    \begin{tabular}{cccc}
    \raggedright
        \begin{overpic}[width=0.242\textwidth]{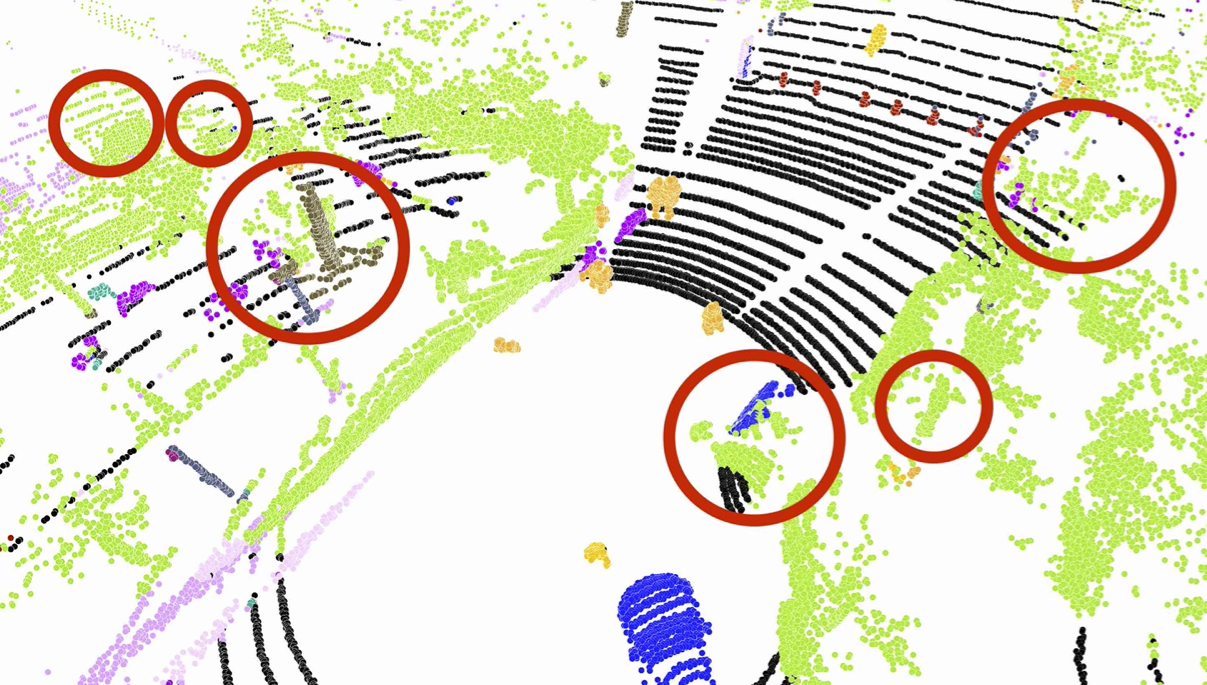}
        \put(40,58){\color{black}\footnotesize \textbf{Source}$^\star$}
        \put(130,58){\color{black}\footnotesize \textbf{\ourmethod-UDA}}
        \put(235,58){\color{black}\footnotesize \textbf{\ourmethod-SSDA}}
        \put(350,58.5){\color{black}\footnotesize \textbf{GT}}
        \end{overpic} &  
        \begin{overpic}[width=0.242\textwidth]{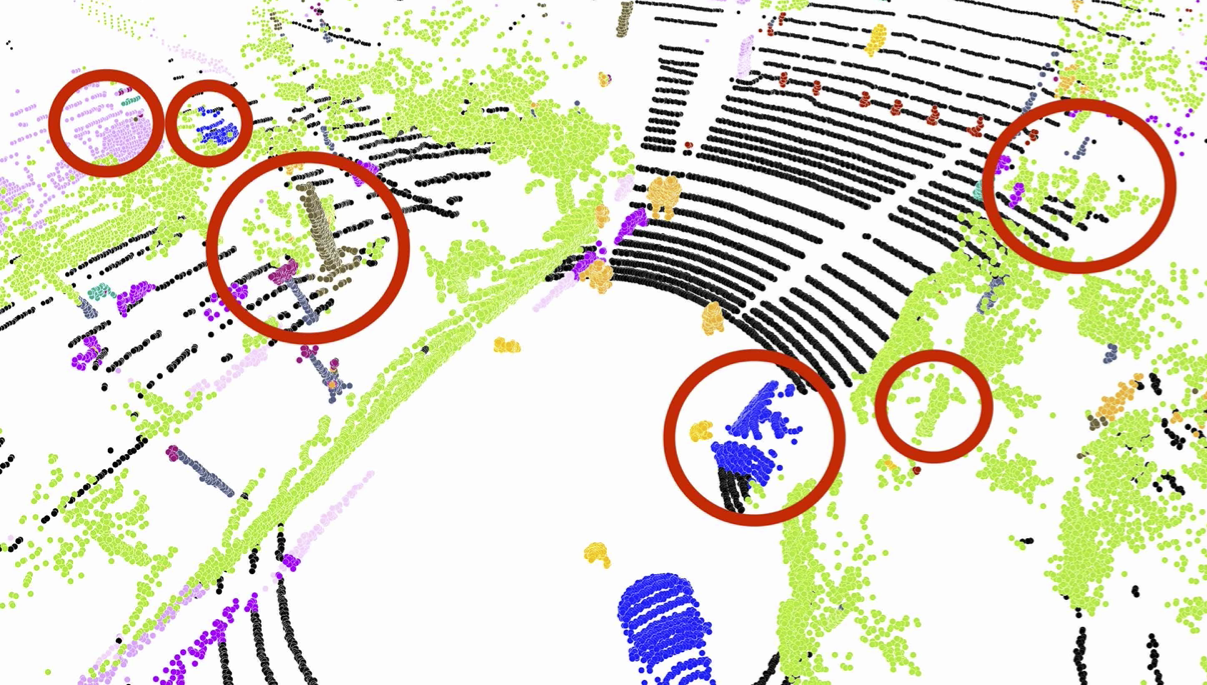}
        \end{overpic} &
        \begin{overpic}[width=0.242\textwidth]{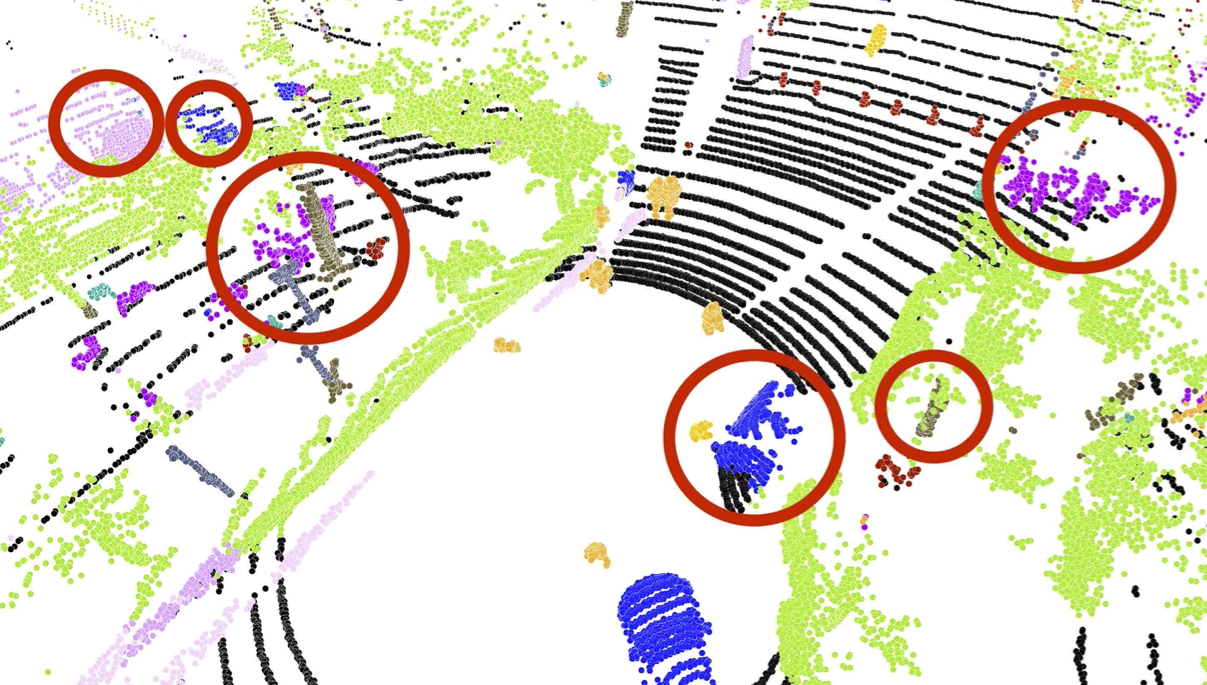}
        \end{overpic} &
        \begin{overpic}[width=0.242\textwidth]{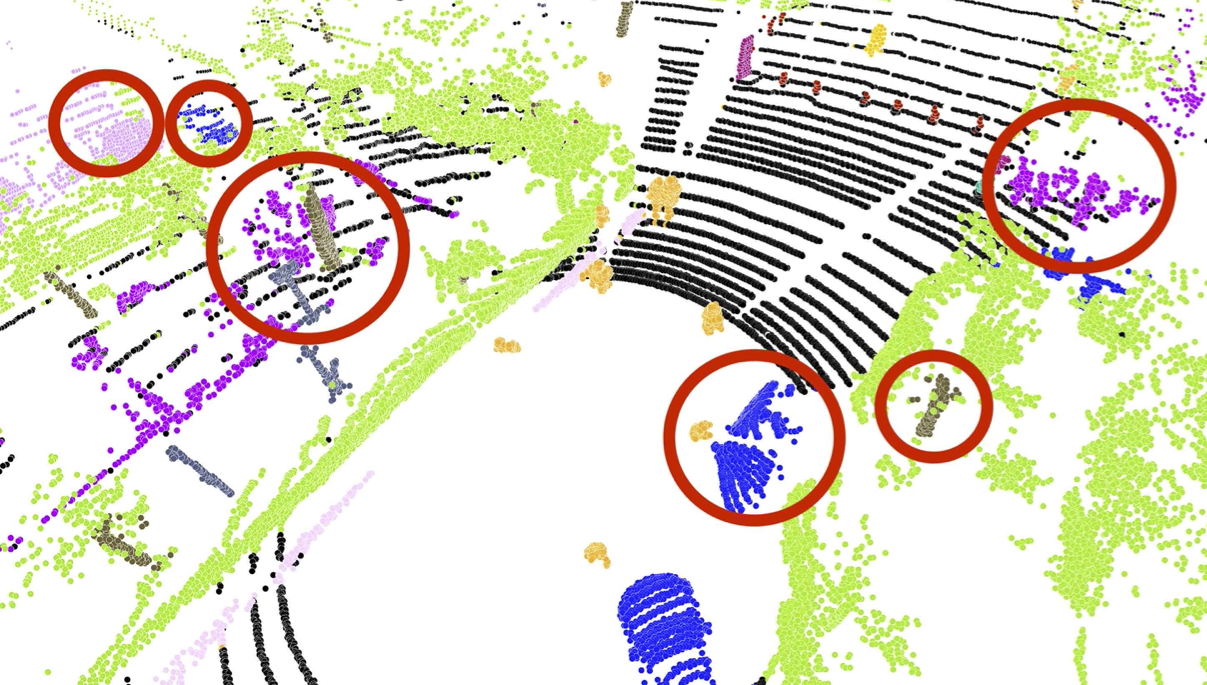}
        \end{overpic}\\
        \begin{overpic}[width=0.242\textwidth]{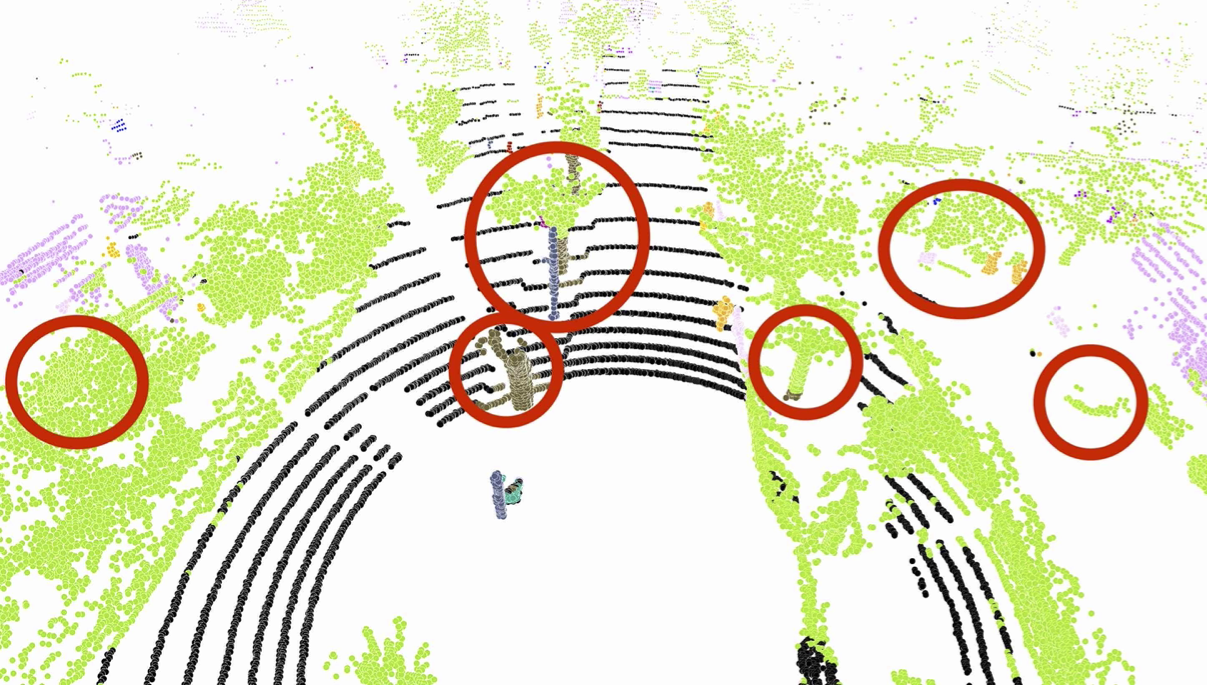}
        \end{overpic} &  
        \begin{overpic}[width=0.242\textwidth]{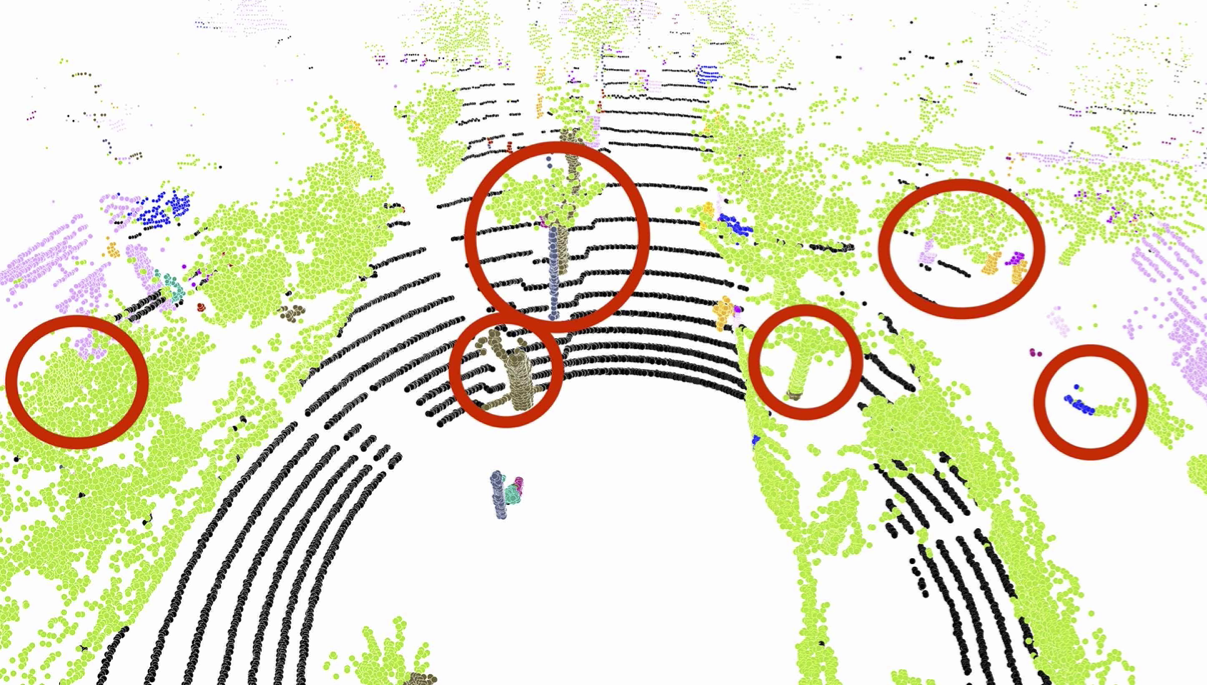}
        \end{overpic} &
        \begin{overpic}[width=0.242\textwidth]{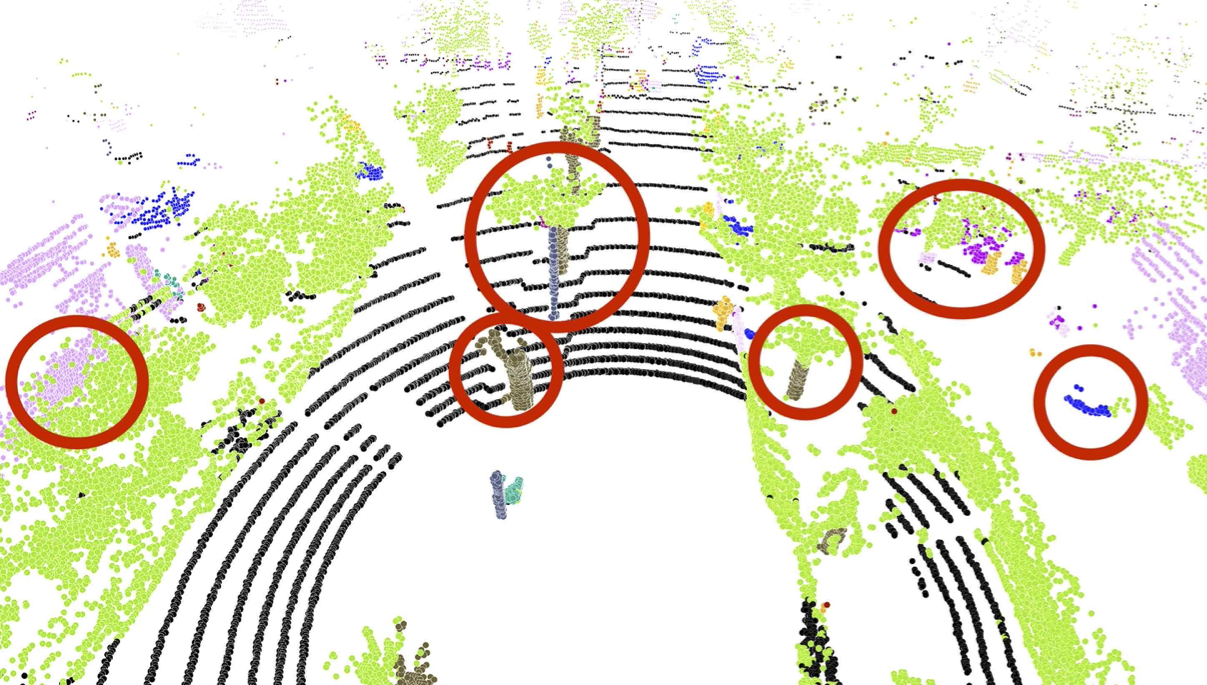}
        \end{overpic} &
        \begin{overpic}[width=0.242\textwidth]{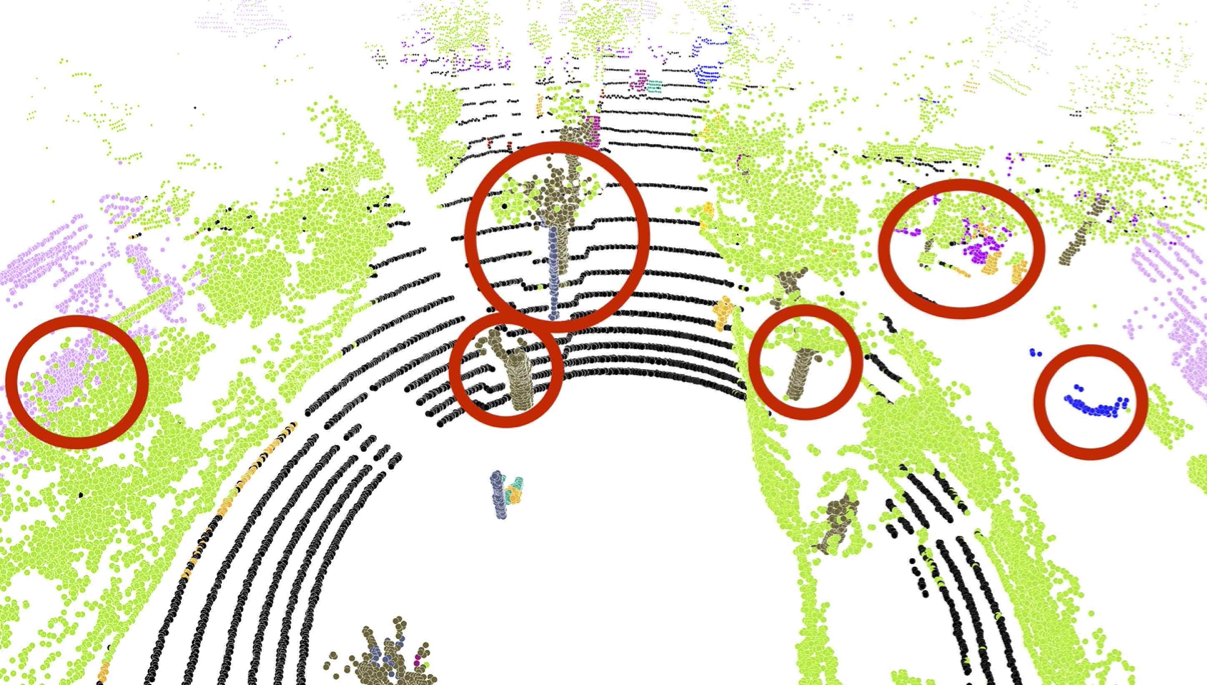}
        \end{overpic}\\
        \begin{overpic}[width=0.242\textwidth]{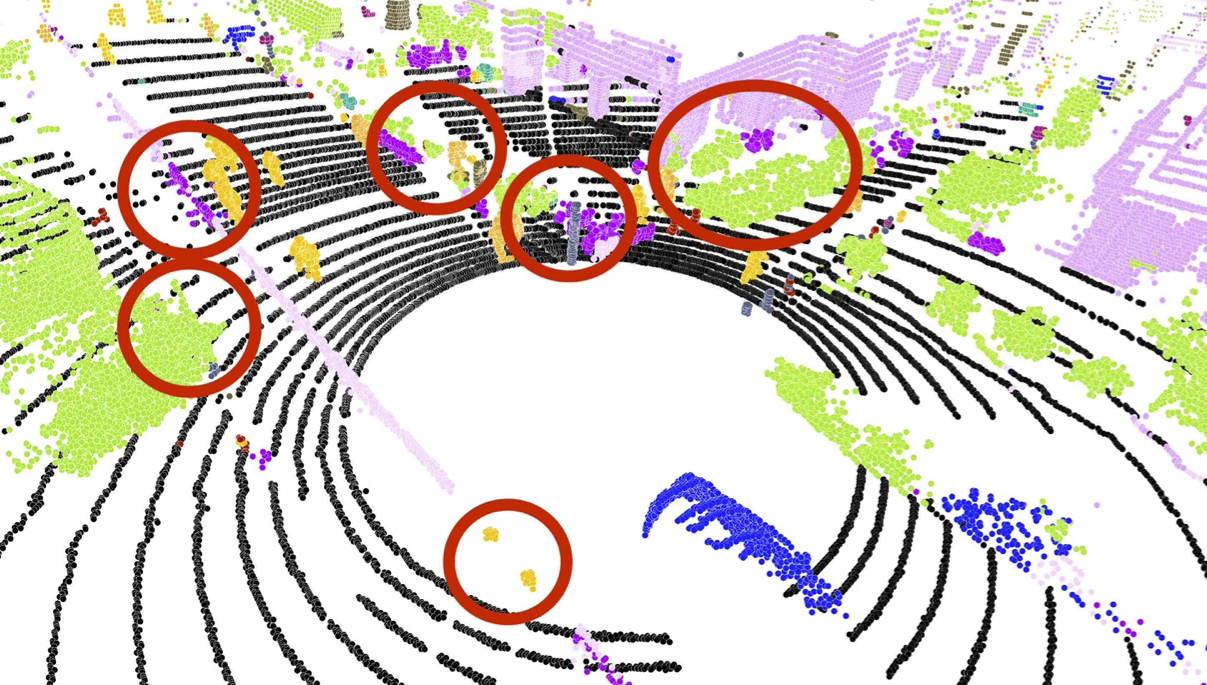}
        \end{overpic} &  
        \begin{overpic}[width=0.242\textwidth]{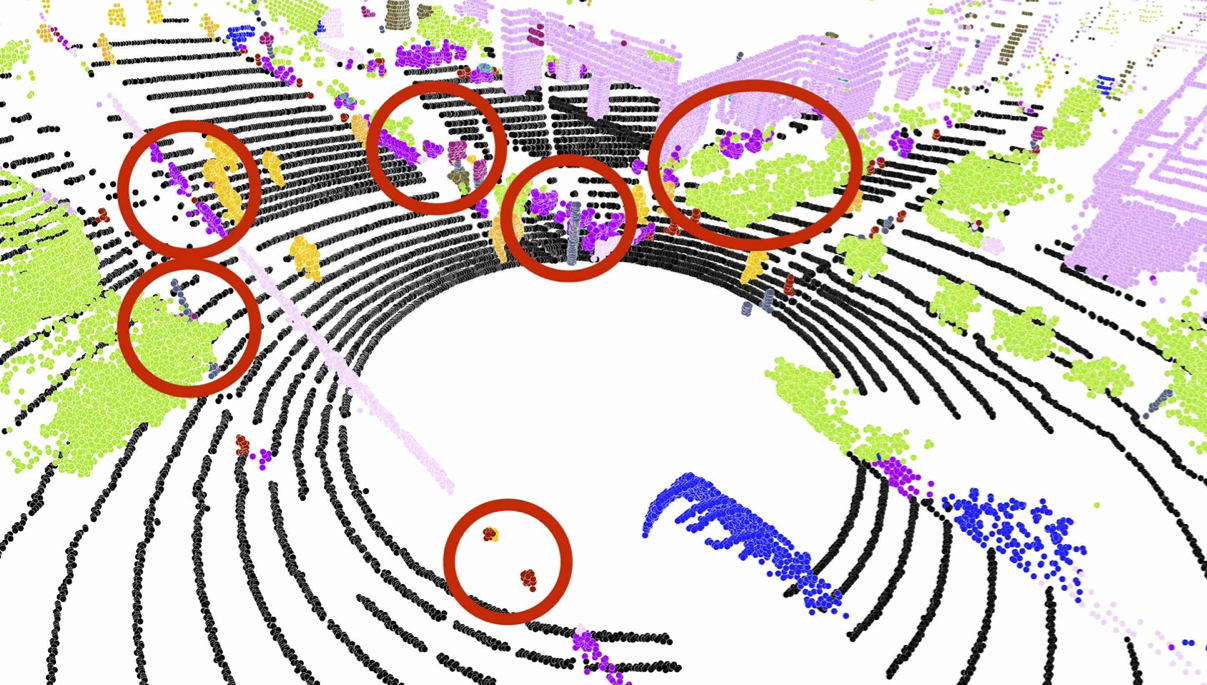}
        \end{overpic} &
        \begin{overpic}[width=0.242\textwidth]{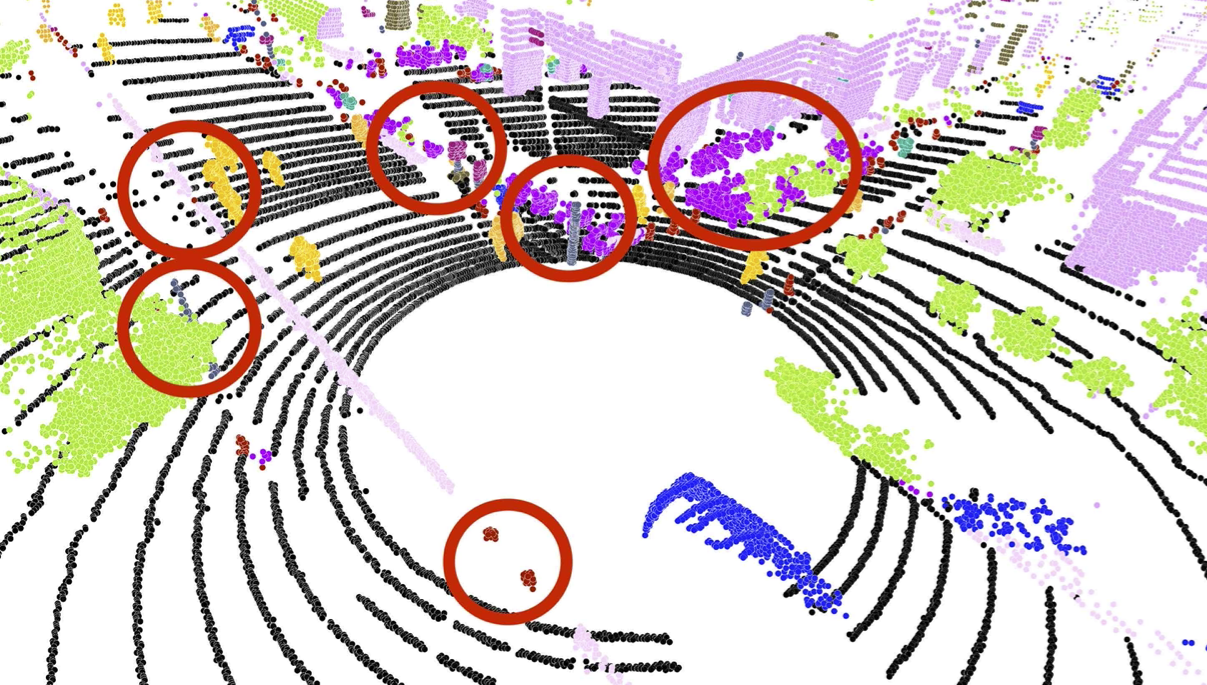}
        \end{overpic} &
        \begin{overpic}[width=0.242\textwidth]{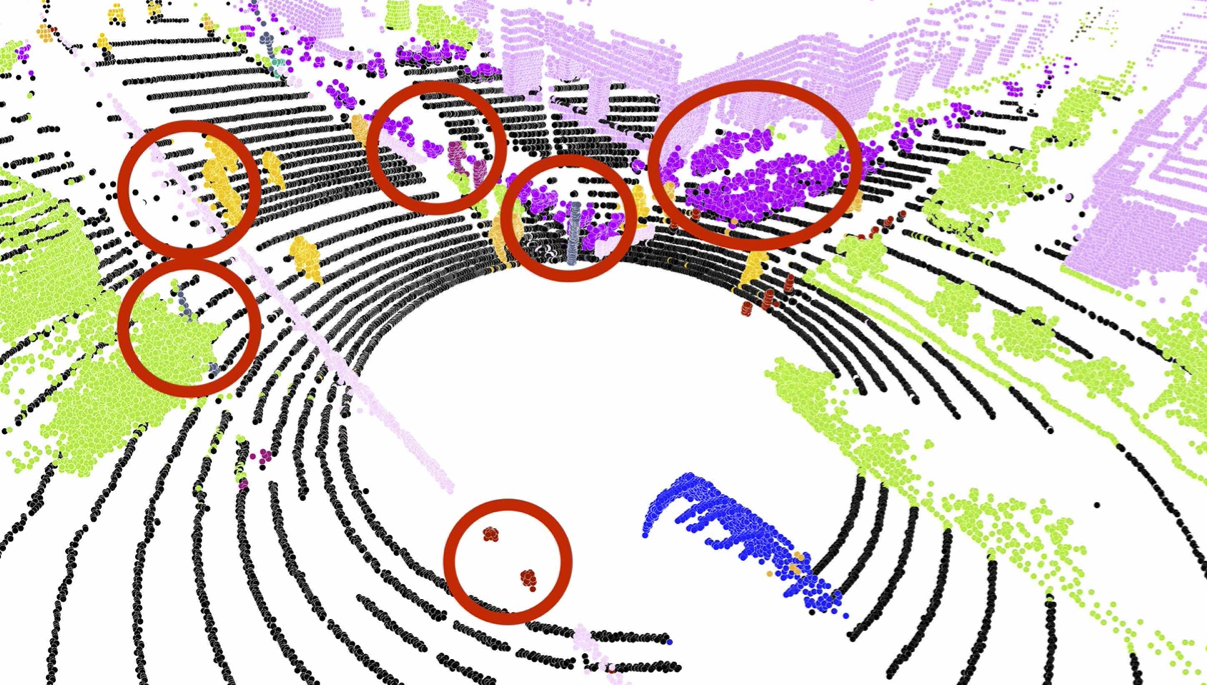}
        \end{overpic}\\
        \multicolumn{4}{c}{
        \begin{overpic}[width=0.99\textwidth]{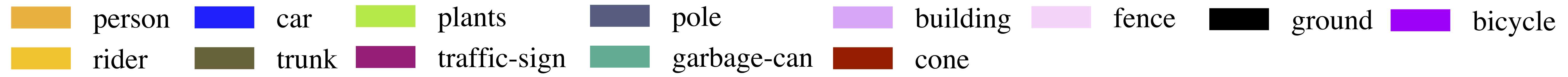}
        \end{overpic}}
    \end{tabular}
    \vspace{-3mm}
    \caption{Results on SynLiDAR $\rightarrow$ SemanticPOSS. \textit{Source}$^\star$ predictions are often wrong and mingled in the same region. After adaptation, \ourmethod-UDA and \ourmethod-SSDA improves segmentation with homogeneous predictions and correctly assigned classes. The red circles highlight regions with interesting results.}
    \label{fig:qualitative_poss}
\end{figure*}

Fig.~\ref{fig:qualitative_poss} shows some domain adaptation results on SynLiDAR $\rightarrow$ SemanticPOSS.
Predictions of \textit{Source}$^\star$ are often incorrect.
\ourmethod-UDA improves the segmentation results with more homogeneous regions and correctly assigned classes, and \ourmethod-SSDA further improves the segmentation quality.

\begin{figure*}[t]
\centering
    \setlength\tabcolsep{1.pt}
    \begin{tabular}{cccc}
    \raggedright
        \begin{overpic}[width=0.242\textwidth]{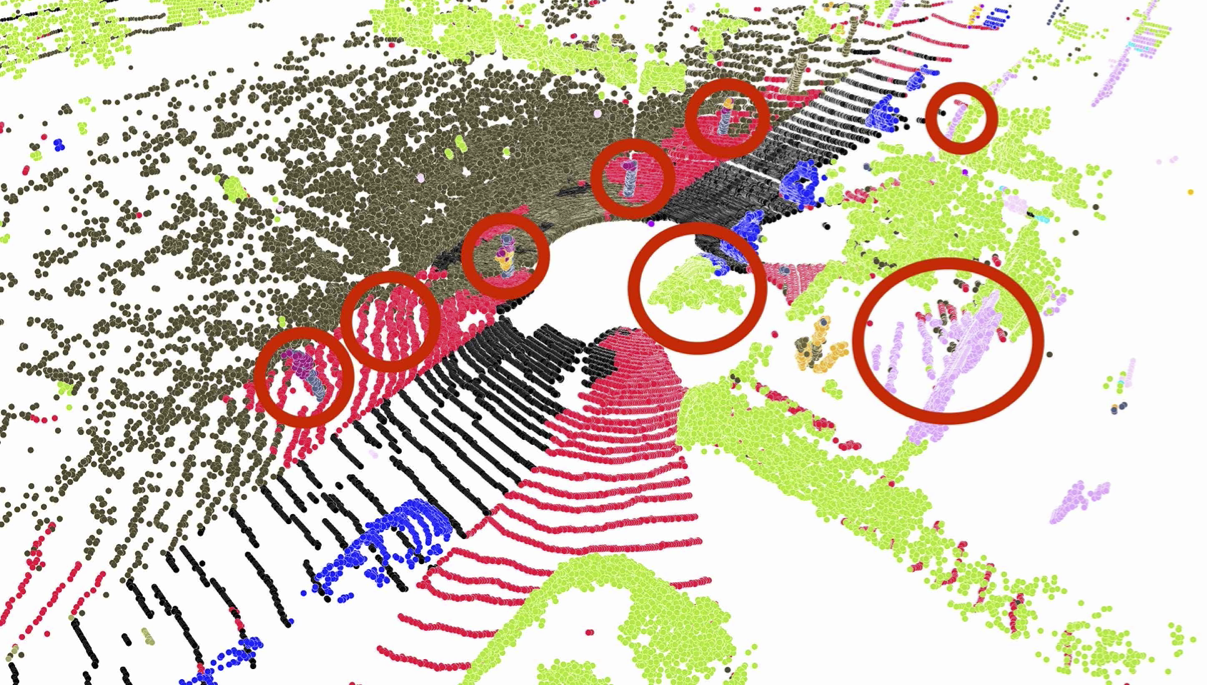}
        \put(40,58){\color{black}\footnotesize \textbf{Source}$^\star$}
        \put(130,58){\color{black}\footnotesize \textbf{\ourmethod-UDA}}
        \put(235,58){\color{black}\footnotesize \textbf{\ourmethod-SSDA}}
        \put(350,58.5){\color{black}\footnotesize \textbf{GT}}
        \end{overpic} &  
        \begin{overpic}[width=0.242\textwidth]{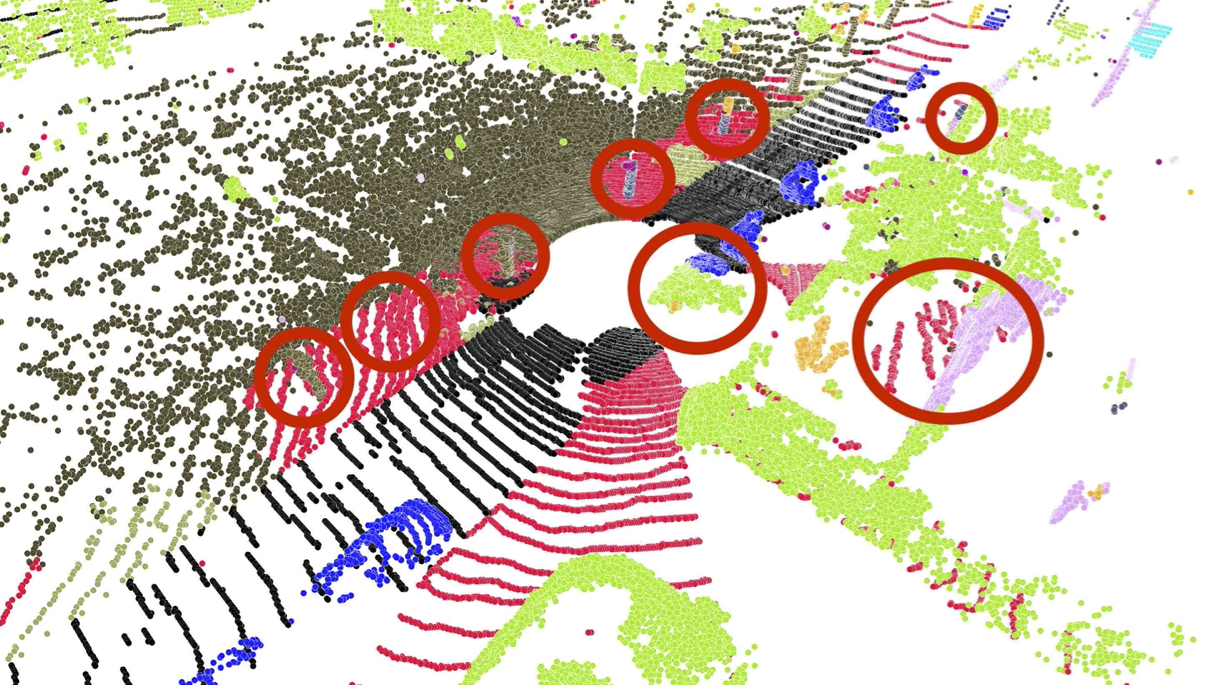}
        \end{overpic} &
        \begin{overpic}[width=0.242\textwidth]{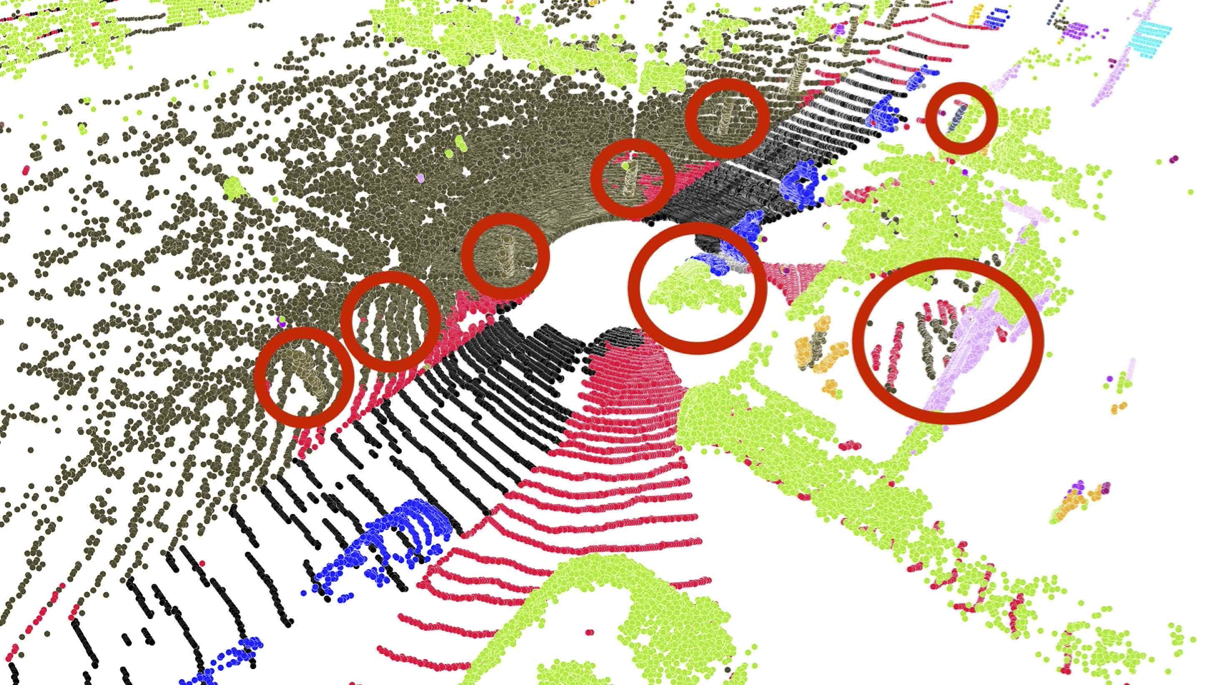}
        \end{overpic} &
        \begin{overpic}[width=0.242\textwidth]{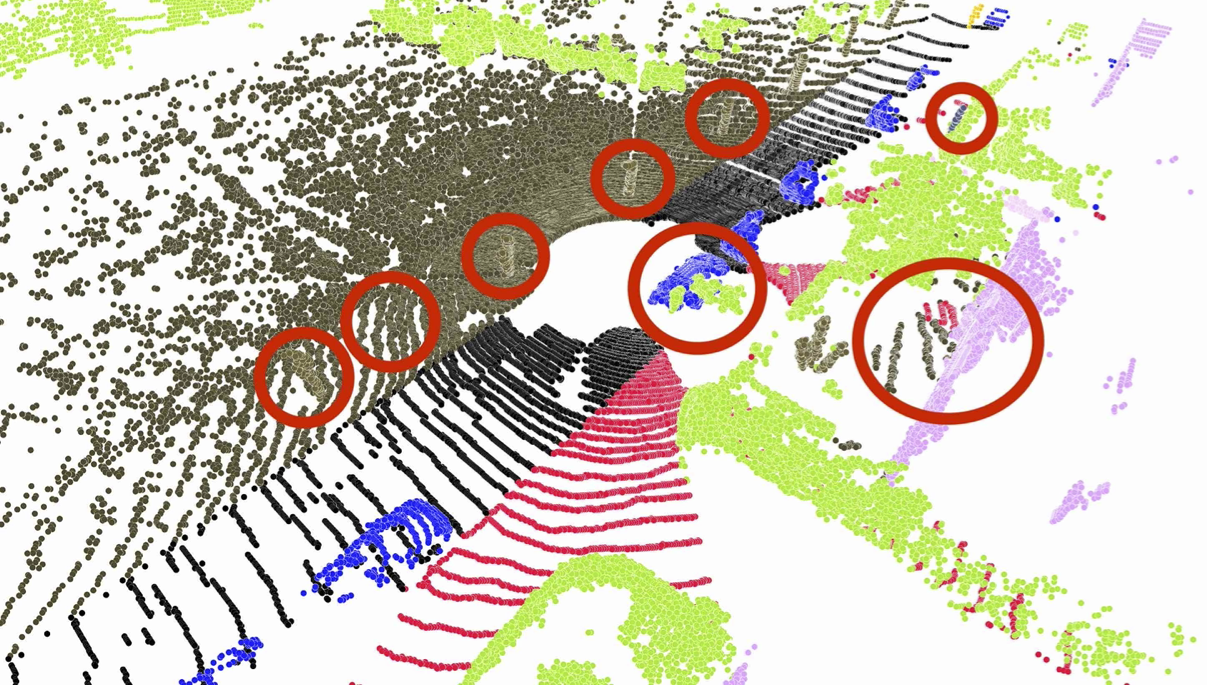}
        \end{overpic}\\
        \begin{overpic}[width=0.242\textwidth]{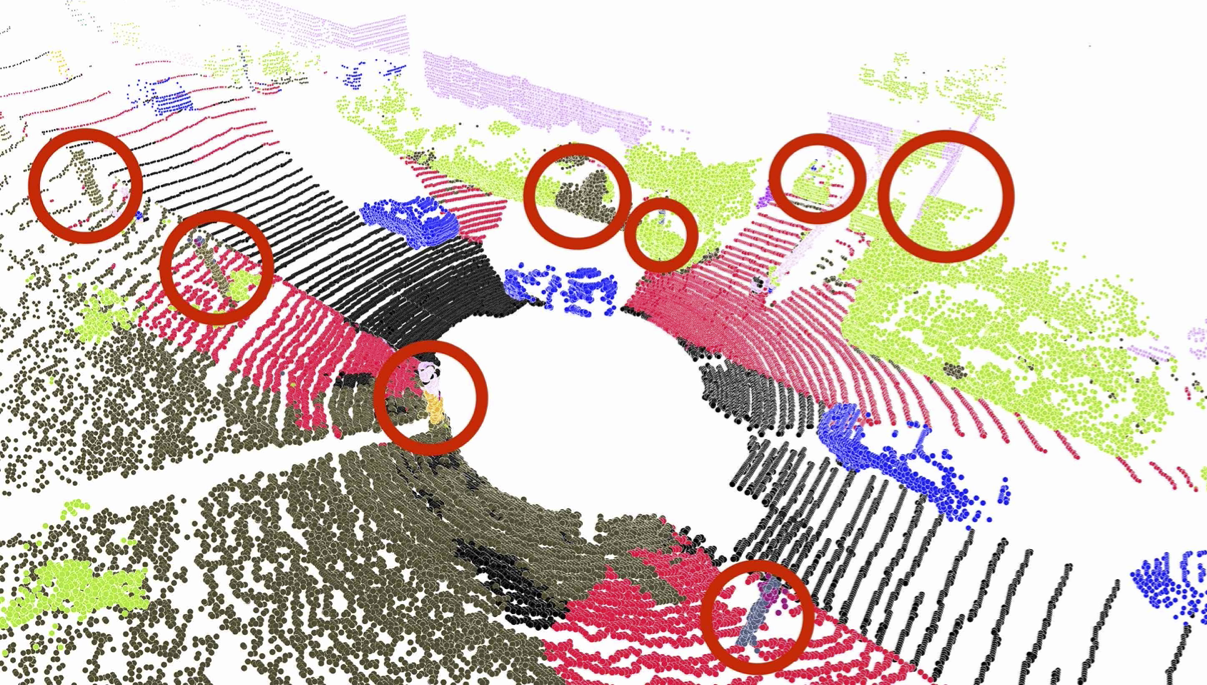}
        \end{overpic} &  
        \begin{overpic}[width=0.242\textwidth]{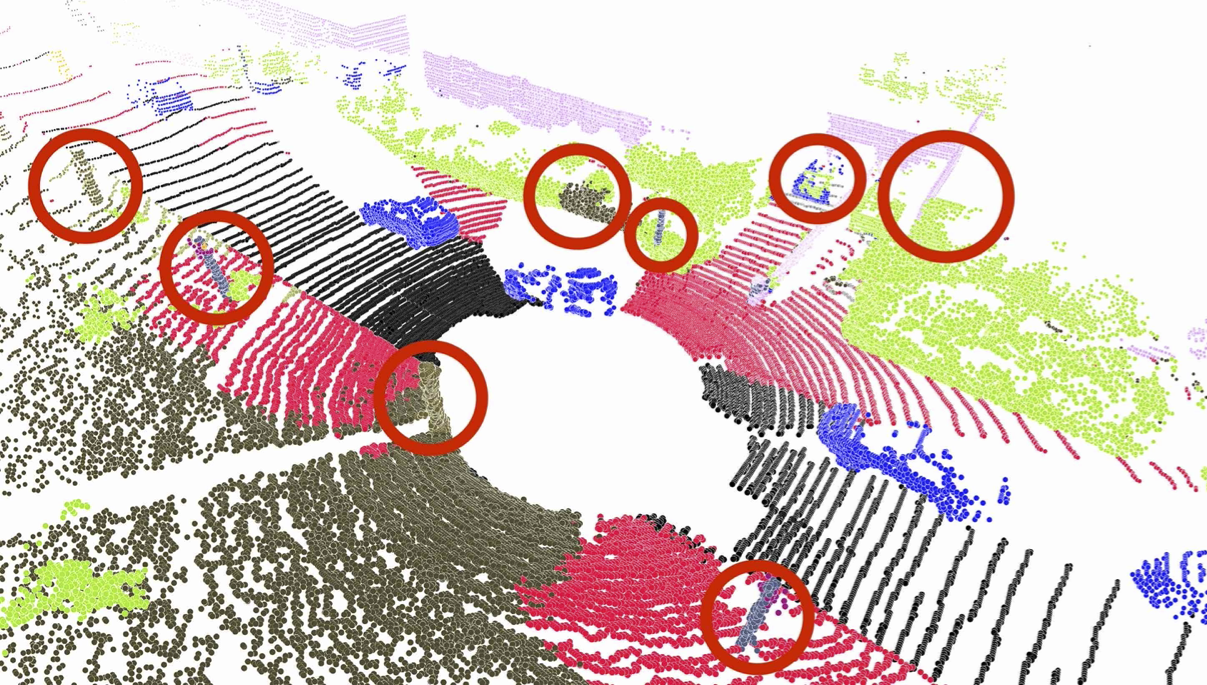}
        \end{overpic} &
        \begin{overpic}[width=0.242\textwidth]{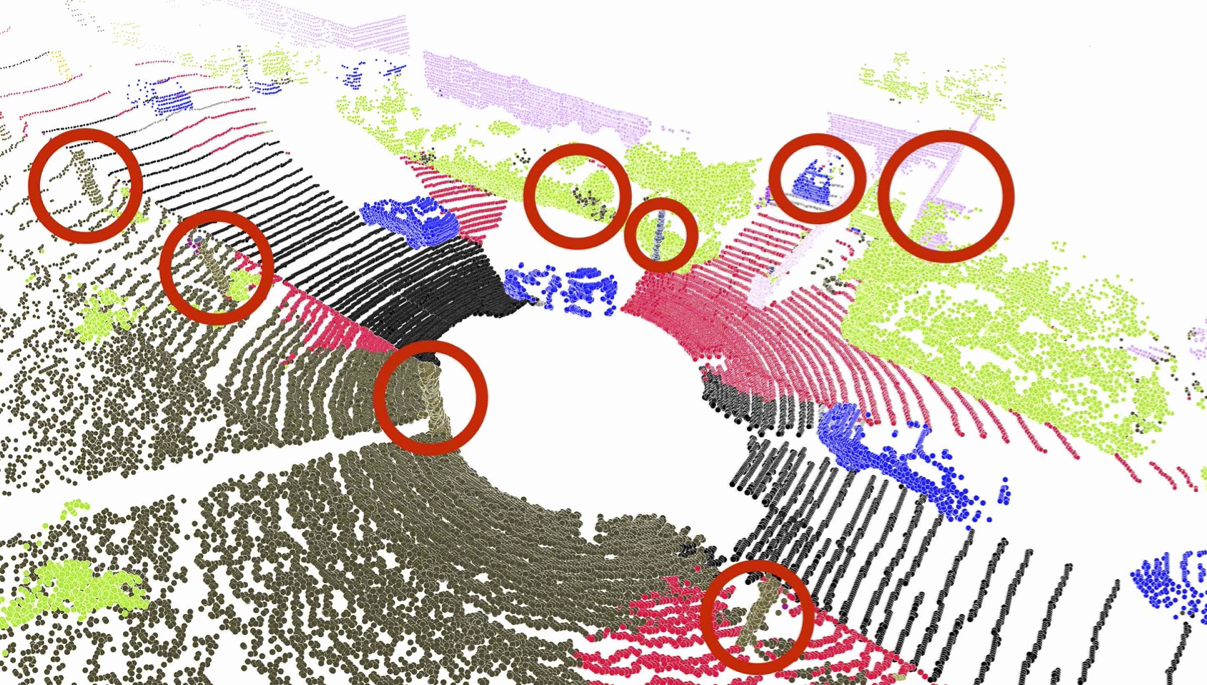}
        \end{overpic} &
        \begin{overpic}[width=0.242\textwidth]{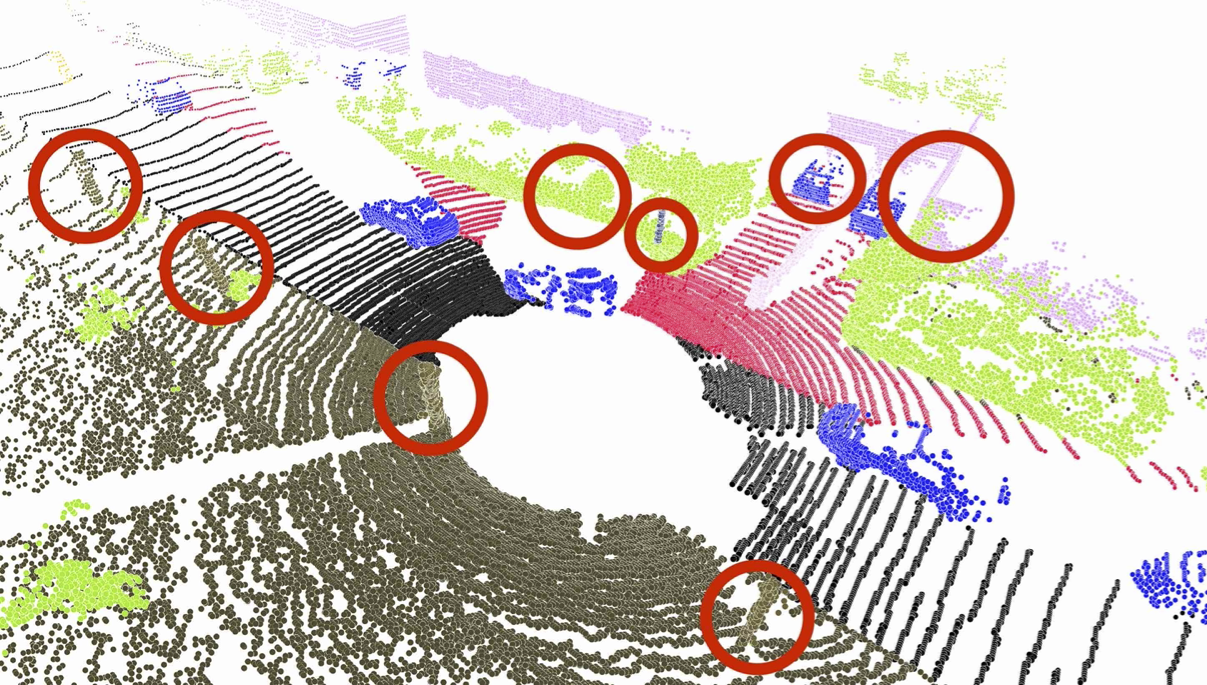}
        \end{overpic}\\
        \begin{overpic}[width=0.242\textwidth]{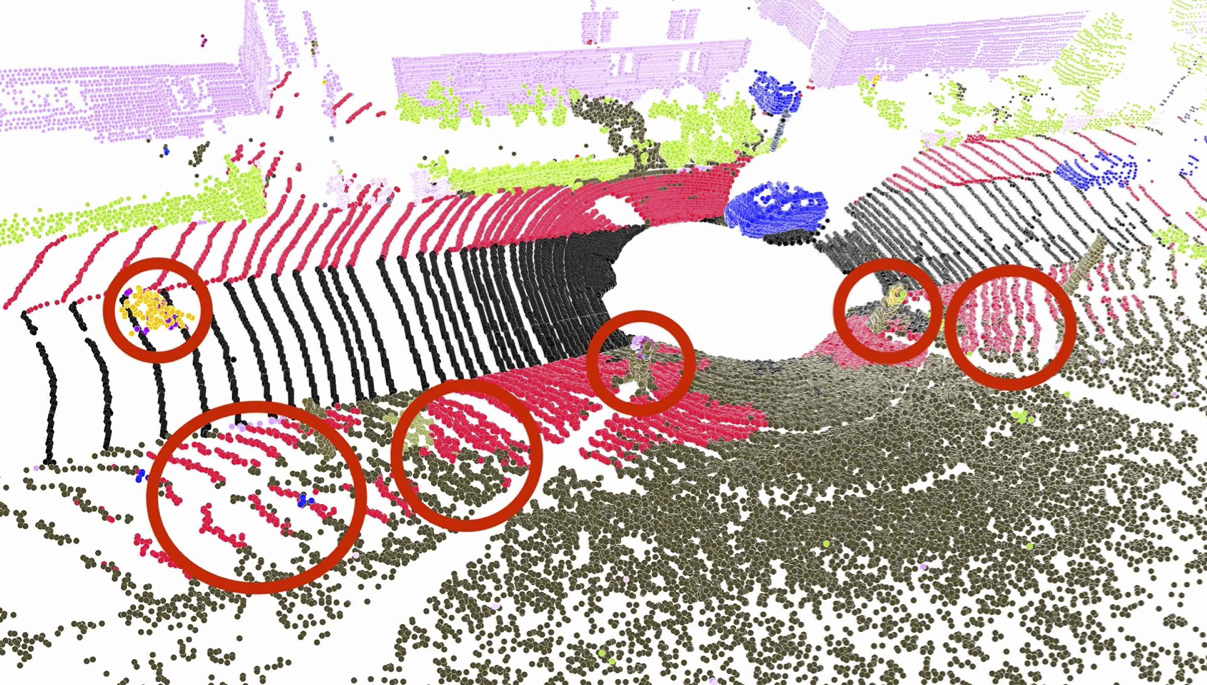}
        \end{overpic} &  
        \begin{overpic}[width=0.242\textwidth]{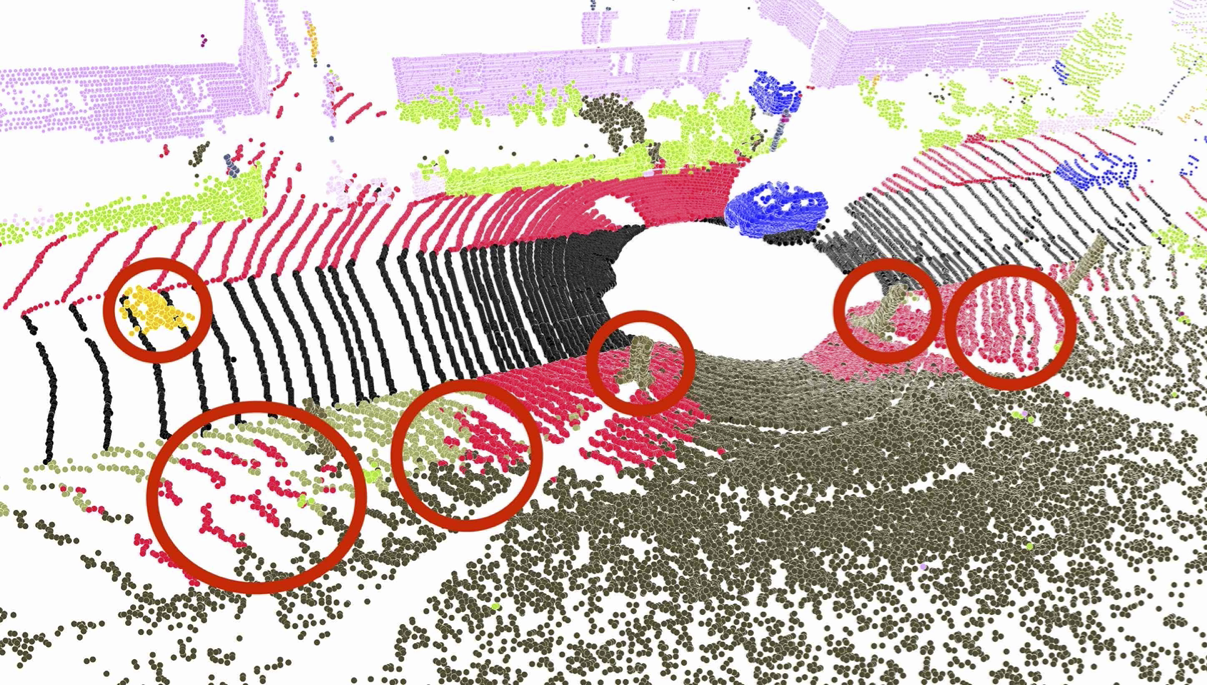}
        \end{overpic} &
        \begin{overpic}[width=0.242\textwidth]{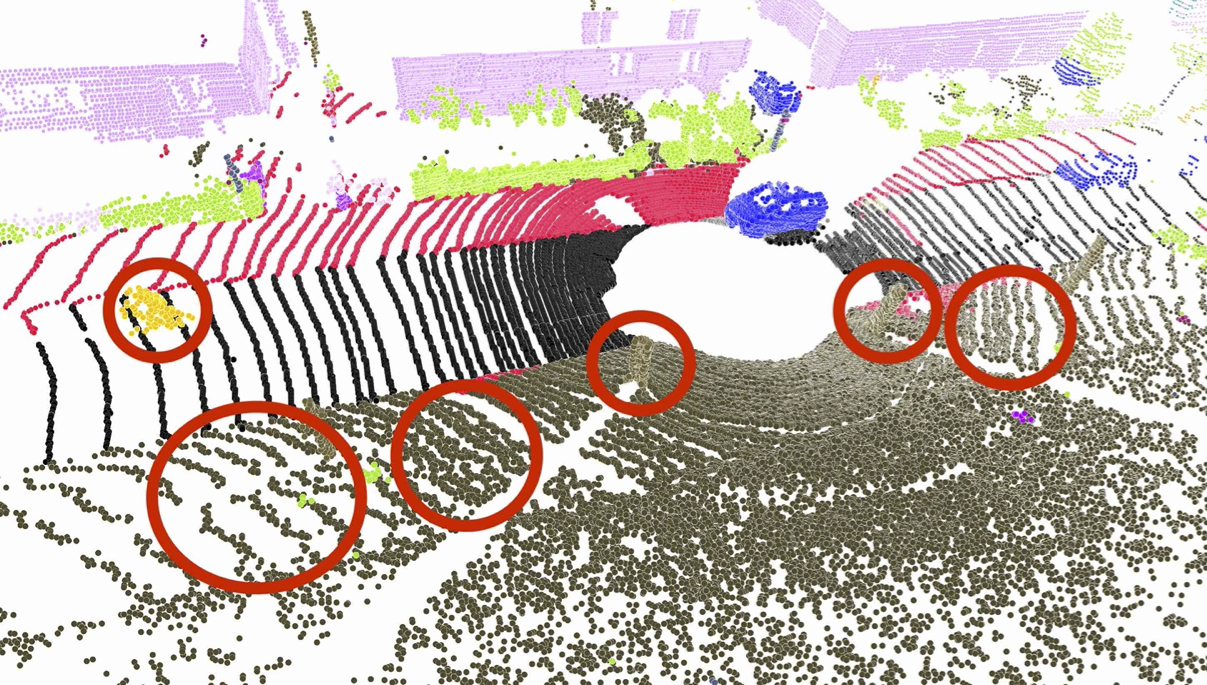}
        \end{overpic} &
        \begin{overpic}[width=0.242\textwidth]{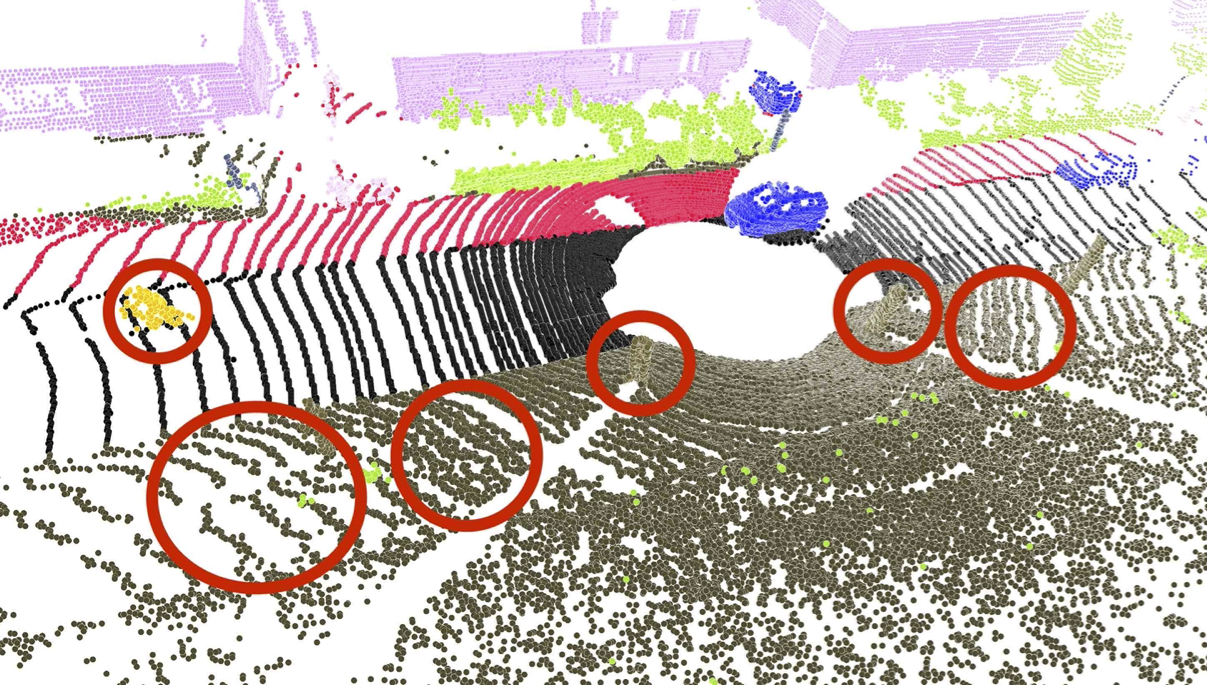}
        \end{overpic}\\
        \multicolumn{4}{c}{
        \begin{overpic}[width=0.98\textwidth]{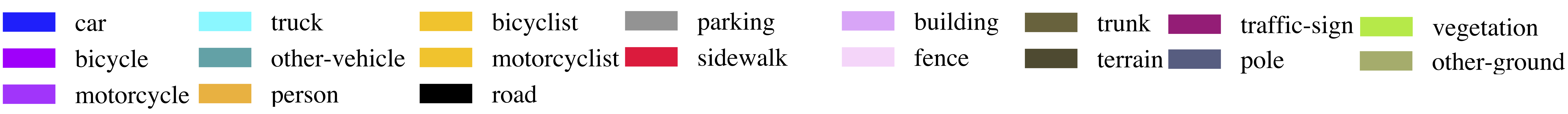}
        \end{overpic}}
    \end{tabular}
    \vspace{-3mm}
    \caption{Results on SynLIDAR $\rightarrow$ SemanticKITTI. \textit{Source}$^\star$ predictions are often wrong and mingled in the same region. 
    After adaptation, \ourmethod-UDA and \ourmethod-SSDA improves segmentation with homogeneous predictions and correctly assigned classes. 
    The red circles highlight regions with interesting results.}
    \label{fig:qualitative_kitti}
\end{figure*}

Fig.~\ref{fig:qualitative_kitti} shows the results on SynLiDAR $\rightarrow$ SemanticKITTI that follows the same result pattern as in Fig.~\ref{fig:qualitative_poss}.
Some classes (e.g.~\textit{car}, \textit{vegetation}, \textit{pole}) greatly improve when \ourmethod-SSDA is used.
An evident increment of performance can be observed from \textit{Source}$^\star$ to \ourmethod-UDA to \ourmethod-SSDA in both the studied domains.
Although the limited amount of target supervision used in \ourmethod-SSDA, these experiments show evidence of the benefits of our SSDA method.

\begin{table}[t]
    \centering
    \caption{Ablation study of the \ourmethod components: mixing strategy ($t\rightarrow s$ and $s\rightarrow t$), compositional mix augmentations (local $h$ and global $r$), mean teacher update ($\beta$) and, weighted class selection in semantic selection ($f$). Each combination is named with a different version (a-h). \textit{Source}$^\star$ performance are added as lower bound and highlighted in gray to facilitate the reading.}
    \label{tab:ablation_components}
    \tabcolsep 5pt
    \resizebox{.75\columnwidth}{!}{%
    \begin{tabular}{c|cc|cc|cc|c}
        \toprule
        \ourmethod & \multicolumn{2}{c|}{mix} & \multicolumn{2}{c|}{augs} & \multicolumn{2}{c|}{others} & \\
         version & $t \rightarrow s$ & $s \rightarrow t$ & $h$ & $r$ & $\beta$ & $f$ & \textbf{mIoU}\\
        \midrule
         \textit{Source}$^\star$ & - & - & - & - & - & - & 21.6\\
        \midrule
        (a) & \ding{51} & & & & & & 31.6\\
        (b) & \ding{51} & & & & \ding{51} & & 31.9\\
        (c) & \ding{51} & \ding{51} &  & & & & 35.0\\
        (d) & \ding{51} & \ding{51} & & &\ding{51} & & 35.4\\
        (e) & \ding{51} & \ding{51} &\ding{51} & &\ding{51} & & 36.8\\
        (f) & \ding{51} & \ding{51} & & \ding{51} &\ding{51} & & 37.3\\
        (g) & \ding{51} & \ding{51} &\ding{51} & \ding{51} & \ding{51} & & 39.0\\
        (h) & \ding{51} & \ding{51} &\ding{51} & \ding{51} & & \ding{51} & 39.1\\
        \midrule
        Full & \ding{51} & \ding{51} & \ding{51} & \ding{51} & \ding{51} & \ding{51} & \textbf{40.4}\\
        \bottomrule
    \end{tabular}
    }
\end{table}

\section{Ablation study}\label{sec:ablations}

We investigate the performance of \ourmethod in both its UDA and SSDA variants by using the SynLiDAR $\rightarrow$ SemanticPOSS setup.
The first three experiments are designed to study \ourmethod in the UDA setting.
In Sec.~\ref{sec:components}, we analyze \ourmethod-UDA components.
In Sec.~\ref{sec:mix_up}, we compare our mixing approach with three recent point cloud mixing strategies, namely, Mix3D~\cite{Nekrasov213DV}, PointCutMix~\cite{zhang2021pointcutmix} and PolarMix~\cite{xiao2022polarmix}.
In Sec.~\ref{sec:noise_pseudo}, we investigate the robustness of \ourmethod to noisy pseudo-labels by changing the confidence threshold $\zeta$ and with different pre-trained models.
In the last experiment (Sec.~\ref{sec:target_supervision}), we analyze \ourmethod in the SSDA setting, comparing our semi-supervised mixing approach with three variations of our approach.

\subsection{Method components}\label{sec:components}

We analyze \ourmethod by organizing its components into three groups: mixing strategies (\textit{mix}), augmentations (\textit{augs}) and other components (\textit{others}).
In the \textit{mix} group, we assess the importance of the mixing strategies ($t\rightarrow s$ and $s\rightarrow t$) used in our compositional mix (Sec.~\ref{sec:compositional_mix}) after semantic selection. 
In the \textit{augs} group, we assess the importance of the local $h$ and global $r$ augmentations that are used in the compositional mix (Sec.~\ref{sec:compositional_mix}). 
In the \textit{others} group, we assess the importance of the mean teacher update ($\beta$) (Sec.~\ref{sec:network_update}) and of the long-tail weighted sampling $f$ (Sec.~\ref{sec:semantic_selection}).
When the $t \rightarrow s$ branch is active, also the pseudo-label filtering $g$ is utilized, while when $f$ is not active, $\alpha=0.5$ source classes are selected randomly.
With different combinations of components, we obtain different versions of \ourmethod which we name \ourmethod (a-h).
The complete version of our method is named \textit{Full}, where all the components are activated. 
The \textit{Source}$^\star$ performance is also added as a reference for the lower bound.
See Tab.~\ref{tab:ablation_components} for the definition of these different versions.

When the $t \rightarrow s$ branch is used, \ourmethod (a) achieves an initial $31.6$ mIoU showing that the $t \rightarrow s$ branch provides a significant adaptation contribution over the \textit{Source}$^\star$. 
When we also use the $s \rightarrow t$ branch and the mean teacher $\beta$, \ourmethod (b-d) further improve performance achieving a $35.4$ mIoU. 
By introducing local and global augmentations in \ourmethod (e-h), we can improve performance up to $39.1$ mIoU. 
The best performance of $40.4$ mIoU is achieved with \ourmethod Full where all the components are activated.

\subsection{Point Cloud Mix}\label{sec:mix_up}
We compare \ourmethod with Mix3D~\cite{Nekrasov213DV}, PointCutMix~\cite{zhang2021pointcutmix} and PolarMix~\cite{xiao2022polarmix} to show the effectiveness of the different mixing designs. 
As per our knowledge, Mix3D and PolarMix are the only mixup strategies designed for 3D semantic segmentation, while PointCutMix and PolarMix are the only strategies for mixing portions of different point clouds.
We implement Mix3D and PointCutMix based on authors descriptions: we concatenate point clouds (random crops for PointCutMix) of the two domains, i.e., $\mathcal{X}^s$ and $\mathcal{X}^t$, as well as their labels and pseudo-labels, i.e., $\mathcal{Y}^s$ and $\hat{\mathcal{Y}}^t$, respectively. 
PolarMix~\cite{xiao2022polarmix} uses our same experimental settings and backbone therefore we consider the results reported in their manuscript.
We refer to these mixing strategies as Mix3D$^\star$, PointCutMix$^\star$ and, PolarMix$^\dagger$.
\ourmethod \textit{double} is our two-branch network with sample mixing.
For a fair comparison, we deactivate the weighted sampling and the mean teacher update.
We keep local and global augmentations activated.

Fig.~\ref{fig:ablations}a shows that Mix3D$^\star$ outperforms the \textit{Source}$^\star$ model, achieving $28.5$ mIoU, followed by PolarMix$^\dagger$ which achieves $30.4$ mIoU.
PointCutMix$^\star$ reaches $31.6$ mIoU, outperforming the previous strategies.
When we use the $t \rightarrow s$ branch alone we can achieve $32.9$ mIoU and when we use the $s \rightarrow t$ branch alone, \ourmethod can further improve the results, achieving $34.8$ mIoU.
This shows that the supervision from the source to target is effective for adaptation on the target domain.
When we use the contribution from both branches simultaneously, \ourmethod achieves the best result with $38.9$ mIoU.

\subsection{Robustness to noisy pseudo-labels}\label{sec:noise_pseudo}

We investigate the robustness of \ourmethod to increasingly noisier pseudo-labels. 
Firstly, we study the effect of different confidence thresholds $\zeta$. 
Secondly, we evaluate different versions of pre-trained models that we use for generating pseudo-labels.

\noindent \textbf{Confidence threshold.} We study the importance of setting the correct confidence threshold $\zeta$ for pseudo-label distillation in $g$ (Sec.~\ref{sec:semantic_selection}). We repeat the experiments with a confidence threshold from $0.65$ to $0.95$ and report the obtained adaptation performance in Fig.~\ref{fig:ablations}b.
\ourmethod is robust to noisy pseudo-labels reaching a $40.2$ mIoU with the low threshold of $0.65$. 
The best adaptation performance of $40.4$ mIoU is achieved with a confidence threshold of $0.85$.
By using a high confidence threshold of $0.95$ performance is affected reaching $39.2$ mIoU. 
With this configuration, too few pseudo-labels are selected to provide an effective contribution for the adaptation.

\begin{figure}
    \centering
    \includegraphics[width=1\columnwidth]{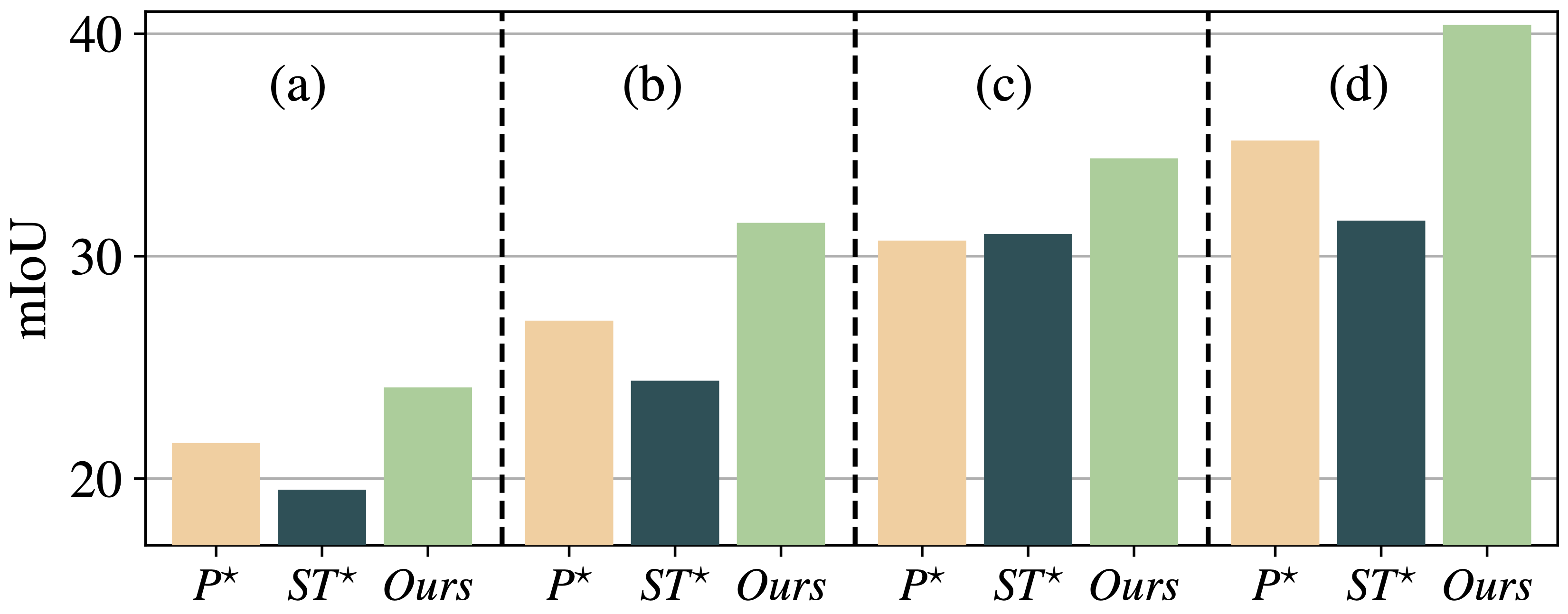}
    \caption{Adaptation results on SynLiDAR$\to$SemanticPOSS with different pre-trained models. We compare the adaptation results of \ourmethod (Ours) with ST$^\star$ starting from different initialization points (P$^\star$) indicated with (a-d).}
    \label{fig:ablation_pseudo}
\end{figure}

\noindent \textbf{Model pre-training.} We quantify the robustness of \ourmethod and ST$^\star$~\cite{zou2019confidence} in response to pseudo-labels generated with different pre-trained models on SynLiDAR and tested on SemanticPOSS. 
In this experiment, we only utilize ST$^\star$ as it is the sole method from those we benchmarked that is based on pseudo-labels.
We denote the pre-trained model as P$^\star$. 
Fig.~\ref{fig:ablation_pseudo} displays its performance at different epochs: (a) 1, (b) 2, (c) 4, and (d) 9.
Unlike \ourmethod, ST$^\star$ proves sensitive to pseudo-labels as it underperforms P$^\star$ in three out of the four cases. 
A plausible explanation for this is that ST$^\star$ refines the pre-trained model using filtered pseudo-labels during adaptation, depending on the quality of pseudo-labels. 
This dependency may cause ST$^\star$ to drift during the adaptation process, thus impacting performance.
Differently, \ourmethod blends source and target (pseudo) labels, producing two intermediate domains with mixed labels. 
In the mixed point clouds, pseudo-labels are integrated with (noise-free) source labels ($t \rightarrow s$) or noise-free selections ($s \rightarrow t$), thus mitigating the negative effects of noisy and imprecise regions.
Furthermore, our application of a teacher-based approach allows us to rely on progressively more precise pseudo-labels, thereby minimizing undesirable drift effects.

\subsection{Mixing target supervision}\label{sec:target_supervision}

\begin{figure*}[t]
\centering
    \setlength\tabcolsep{1.pt}
    \begin{tabular}{ccc}
    \raggedright
        \begin{overpic}[width=0.335\textwidth]{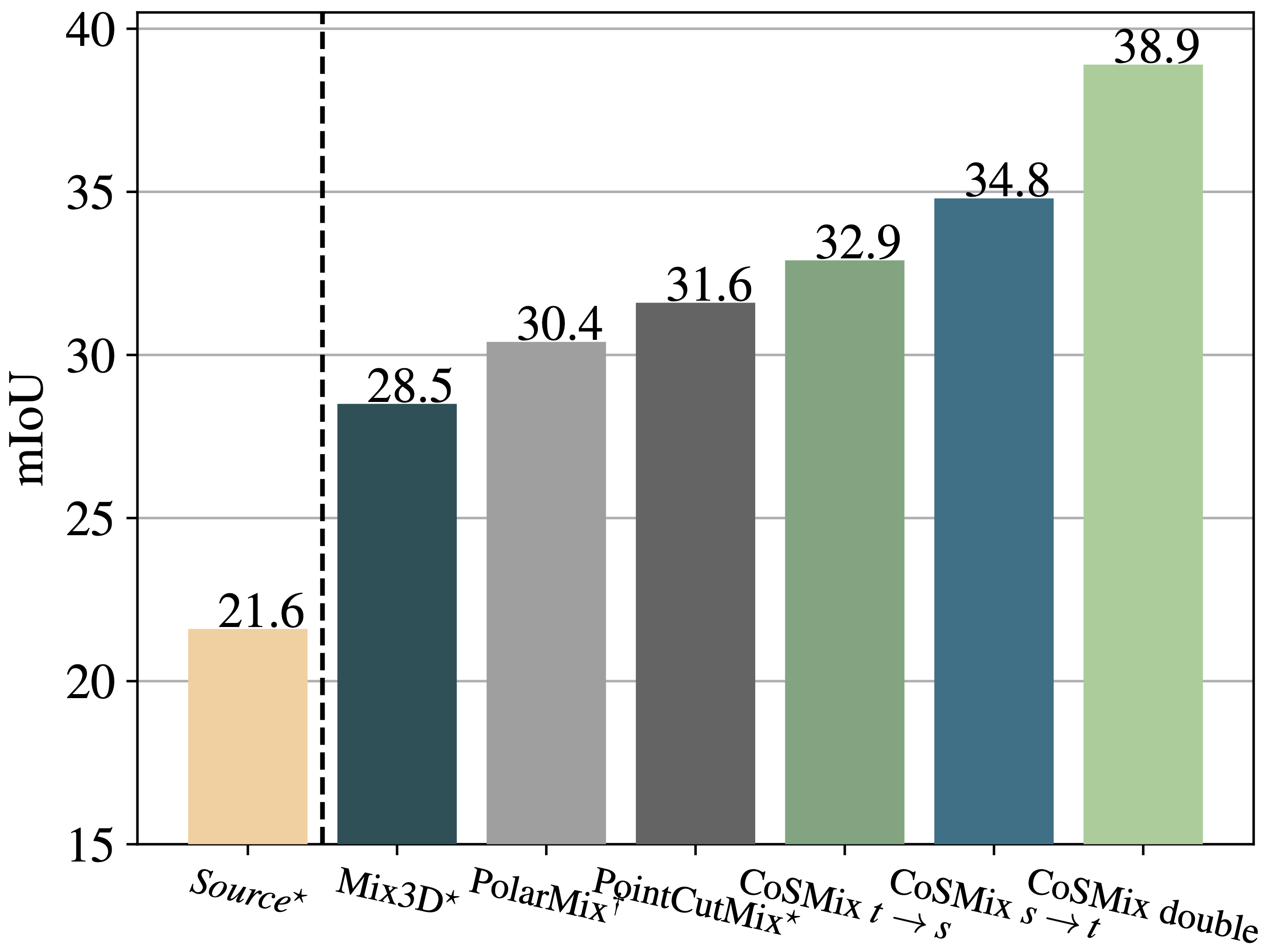}
        \put(55,-5){\color{black}\footnotesize a)}
        \put(150,-5){\color{black}\footnotesize b)}
        \put(250,-5){\color{black}\footnotesize c)}
        \end{overpic} &  
        \begin{overpic}[width=0.337\textwidth]{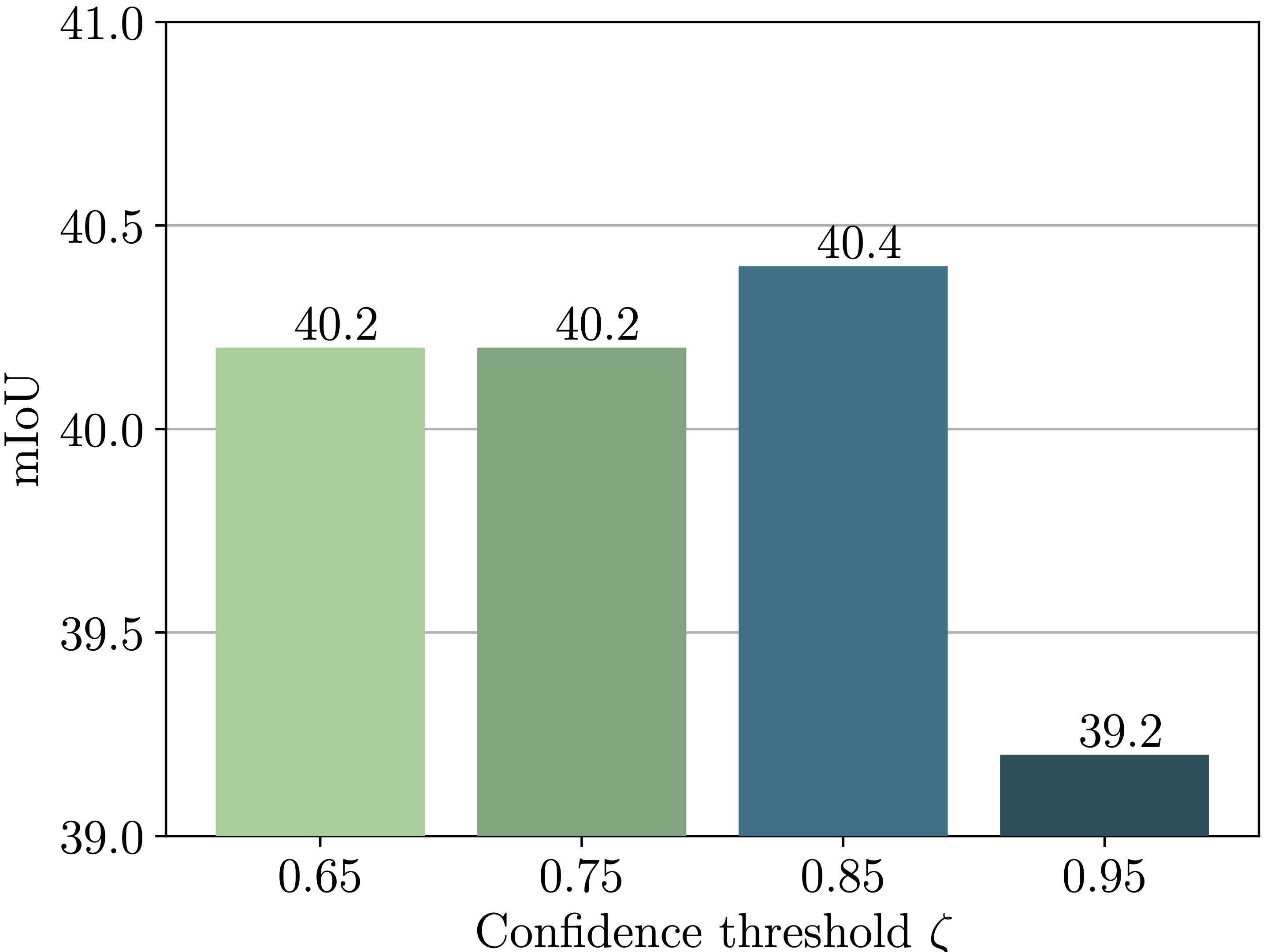}
        \end{overpic} &
        \begin{overpic}[width=0.329\textwidth]{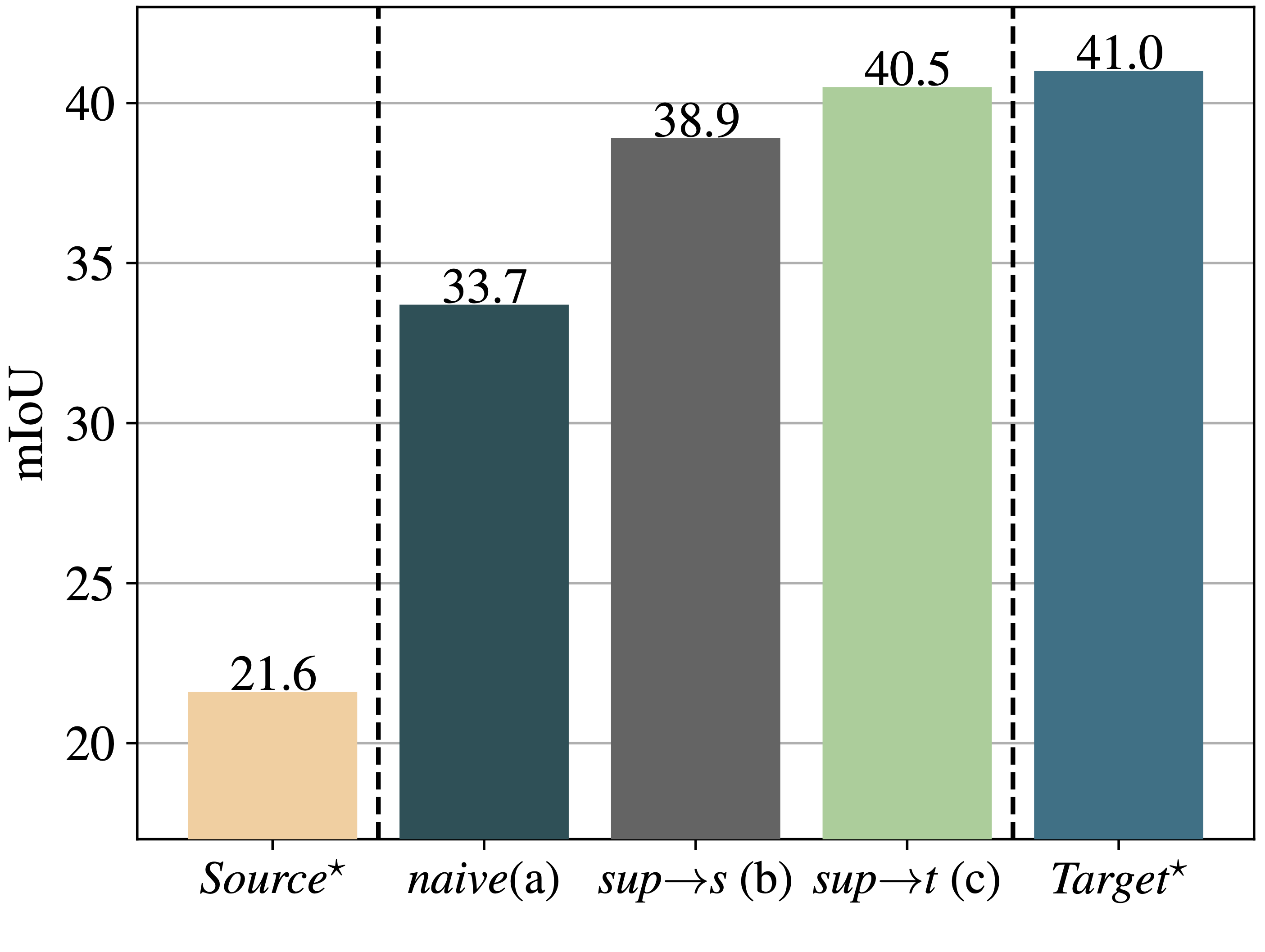}
        \end{overpic}
    \end{tabular}
    \caption{\textbf{a)} Comparison of the adaptation performance with different point cloud mix up strategies. Compared to the recent mixing strategies Mix3D~\cite{Nekrasov213DV}, PointCutMix~\cite{zhang2021pointcutmix} and, PolarMix~\cite{xiao2022polarmix}, our mixing strategy and its variations achieve superior performance. 
    \textbf{b)} Comparison of the adaptation performance on confidence threshold values. Adaptation results show that $\zeta$ should be set such that to achieve a trade-off between pseudo-label correctness and object completeness. \textbf{c)} Comparison of the SSDA performance with different mixing strategies: 
    optimization without mix (naive), single branch mixing with source point clouds ($sup\rightarrow s$), 
    single branch mixing with unsupervised target point clouds ($sup\rightarrow t$). Each variation is named with a different version (a-c). 
    In all the experiments, \textit{Source}$^\star$ and \textit{Target}$^\star$ performance is the lower and upper bound.}
    \label{fig:ablations}
\end{figure*}

We compare \ourmethod-SSDA to three alternative mixing strategies: naive, $sup\rightarrow s$ and $sup\rightarrow t$. In Fig.~\ref{fig:ablations}c, we name each strategy as \ourmethod-SSDA (a-c). 
In version \ourmethod-SSDA (a), we apply \ourmethod-UDA without mixing $\mathcal{T}_\textsf{L}$ in $\mathcal{X}^{s \rightarrow t}$ and $\mathcal{X}^{t \rightarrow s}$. Dice segmentation loss is applied separately on $\mathcal{T}_\textsf{L}$ and averaged with our total objective loss in Eq.~\ref{eq:total_loss}.
In the single branch mixing with source point clouds ($sup\rightarrow s$) and with target point clouds ($sup\rightarrow t$), versions (b-c), we apply only the upper or lower branch of \ourmethod-SSDA, respectively. Full is our proposed double branched \ourmethod-SSDA.

\ourmethod-SSDA (a) approach reaches $33.7$ mIoU, which shows that traditional training by using labeled target points as is leads to inferior performance than using our SSDA approach.
Both the single branch mixing strategies achieve better performance with $38.9$ mIoU and $40.5$ mIoU for $sup\rightarrow s$ and $sup\rightarrow t$, respectively.
The version (b) shows that the mixed target modality with noise-free annotations helps in reducing the domain shift. 
The version (c) suggests that the addition of target noise-free labels helps us in achieving higher performance. However, both the single branch approaches are not sufficient to outperform the Full mixing strategy.

\section{Conclusions}\label{sec:conclusions}

We introduced the first method for domain adaptation in 3D semantic segmentation, featuring a novel 3D point cloud mixing strategy that harnesses both semantic and structural information simultaneously. 
We developed two variations of our approach: one for unsupervised adaptation (\ourmethod-UDA) and another for semi-supervised adaptation (\ourmethod-SSDA).
We performed comprehensive evaluations in both synthetic-to-real and real-to-real contexts within UDA and SSDA settings, utilizing large-scale, publicly available LiDAR datasets. Experimental results demonstrated that our approach significantly surpasses current state-of-the-art methods in both contexts.
Moreover, detailed analyses underscored the significance of each component within \ourmethod, confirming that our mixing strategy effectively addresses the issue of domain shift in 3D LiDAR segmentation.
A primary limitation of \ourmethod is its reliance on pseudo-labels, making the quality of the initial warm-up model on the source domain crucial to the adaptation performance on the target domain.
An alternative approach could involve the implementation of self-supervised learning in lieu of using source data. 
Future avenues for research might encompass the incorporation of self-supervised learning tasks, domain generalization, extending \ourmethod to source-free adaptation tasks, and its application to 3D object detection.


\ifCLASSOPTIONcompsoc
  \section*{Acknowledgments}
\else
  \section*{Acknowledgment}
\fi

This work was partially supported by OSRAM GmbH, by the MUR PNRR project FAIR (PE00000013) funded by the NextGenerationEU, by the EU project FEROX Grant Agreement no 101070440, by the PRIN PREVUE (Prot. 2017N2RK7K), the EU ISFP PROTECTOR (101034216) and the EU H2020 MARVEL (957337) projects and, it was carried out in the Vision and Learning joint laboratory of FBK and UNITN.


\bibliographystyle{IEEEtran}
\bibliography{egbib}

\vspace{-1.3cm}
\begin{IEEEbiography}
[{\includegraphics[width=1in,height=1.25in,clip,keepaspectratio]{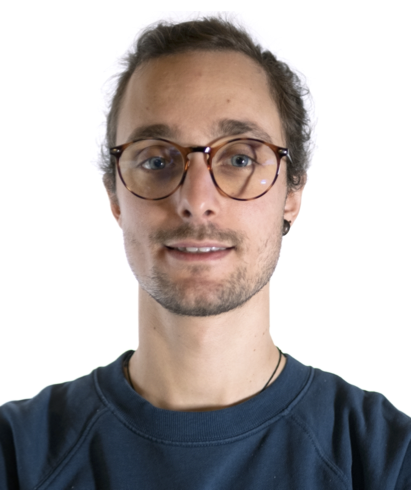}}]{Cristiano Saltori}
is a PhD candidate in Computer Vision at the University of Trento, Italy. Before, he received his Infomation and Communication Engineering Master's degree from the University of Trento. His research interests include deep learning, 3D scene understanding, perception and domain adaptation. 
\end{IEEEbiography}
\vspace{-1.3cm}
\begin{IEEEbiography}
[{\includegraphics[width=1in,height=1.25in,clip,keepaspectratio]{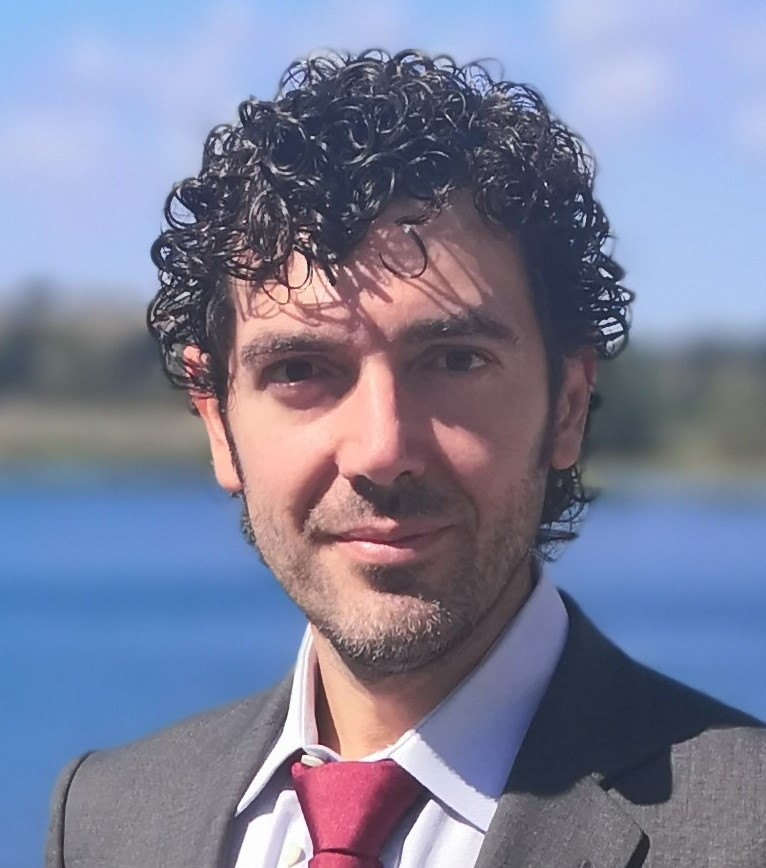}}]{Fabio Galasso} heads the Perception and Intelligence Lab (PINlab) at the Dept.\ of Computer Science, Sapienza University of Rome (Italy). His research interests include distributed and multi-agent intelligent systems, perception (detection, recognition, re-identification, forecasting) and general intelligence (reasoning, meta-learning, domain adaptation), within sustainable and interpretable AI frameworks.
\end{IEEEbiography}
\vspace{-1.3cm}
\begin{IEEEbiography}[{\includegraphics[width=1in,height=1.25in,keepaspectratio]{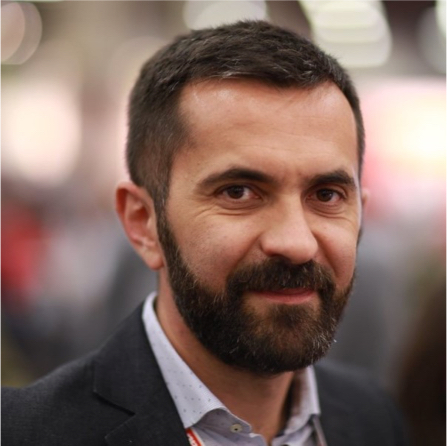}}]{Giuseppe Fiameni} is a Data Scientist at NVIDIA where he oversees the AI Technology Center in Italy, a collaboration among NVIDIA, CINI and CINECA to accelerate academic research in the field of Artificial Intelligence through collaboration projects. He has been working as HPC specialist and consultant at CINECA, the largest HPC facility in Italy, providing support for large-scale data analytics workloads.
\end{IEEEbiography}
\vspace{-1.3cm}
\begin{IEEEbiography}[{\includegraphics[width=1in,height=1.25in,keepaspectratio]{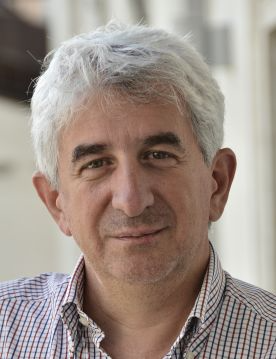}}]{Nicu Sebe} is Professor with the University of
Trento, Italy, leading the research in the areas of
multimedia information retrieval and human behavior
understanding. He was the General Co-
Chair of ACM Multimedia 2013 and 2022, and the Program
Chair of ACM Multimedia 2007 and 2011, ECCV
2016, ICCV 2017 and ICPR 2020. He is a fellow
of the International Association for Pattern
Recognition.
\end{IEEEbiography}
\vspace{-1.3cm}
\begin{IEEEbiography}[{\includegraphics[width=1in,keepaspectratio]{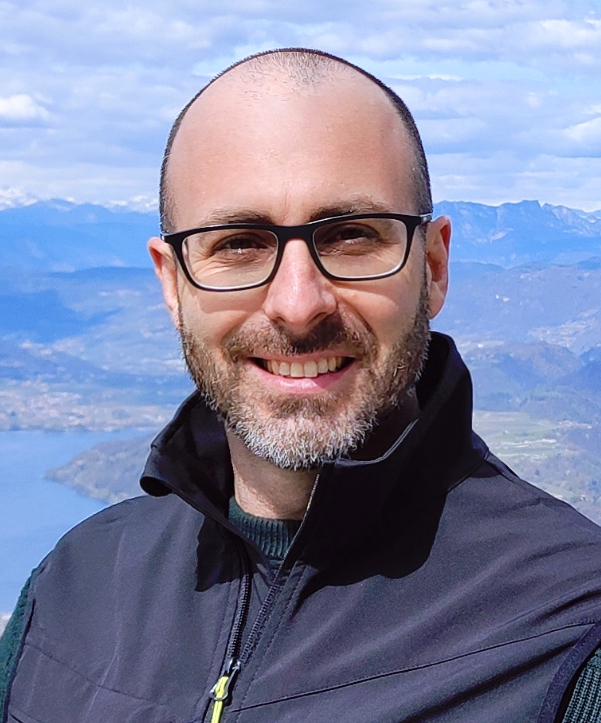}}]{Fabio Poiesi}
is a Researcher in the Technologies of Vision (TeV) Lab at Fondazione Bruno Kessler in Trento, Italy.
He received his Ph.D.~degree from Queen Mary University of London (UK). 
He was a Postdoctoral Researcher in Queen Mary University of London before moving to TeV.
His research interests are 3D scene understanding, object detection and tracking, and extended reality.
\end{IEEEbiography}
\vspace{-1.3cm}
\begin{IEEEbiography}[{\includegraphics[width=1in,height=1.25in,clip,keepaspectratio]{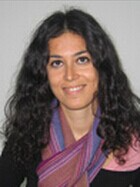}}]
{Elisa Ricci} is an associate
professor at the University of Trento and a head of research unit at Fondazione Bruno Kessler. She
received the Honorable mention award at ICCV
2021 and the Best Paper Award at ACM MM 2021. Her research interests are
mainly in the areas of computer vision and
deep learning. She is an ELLIS fellow. 
\end{IEEEbiography}
\end{document}